\RequirePackage[hyphens]{url}
\documentclass[accepted]{uai2022} 
\usepackage{xr-hyper}
\usepackage{hyperref}
\usepackage[dvipsnames]{xcolor}
\definecolor{c1}{RGB}{41,140,190}

\usepackage[american]{babel}
\usepackage{microtype}
\usepackage{graphicx}
\usepackage{booktabs}
\usepackage{algorithmic}
\usepackage{algorithm}
\usepackage{natbib} 
\usepackage{mathtools} 
\usepackage{booktabs} 
\usepackage{tikz}

\usepackage{amsmath}
\usepackage{amssymb}
\usepackage{amsfonts}
\usepackage{amsthm}
\usepackage{bm}
\newcommand{\mbf}[1]{{\boldsymbol{\mathbf{#1}}}}
\renewcommand{\bm}{\mbf}

\newcommand{\norm}[1]{\left\lVert#1\right\rVert}
\newcommand{\set}[1]{\{#1\}}

\DeclareMathOperator{\tr}{Tr}

\usepackage{nicefrac}
\usepackage{comment}
\usepackage{subcaption}

\title{Low-Precision Arithmetic for Fast Gaussian Processes}

\author[1,*]{Wesley J. Maddox}
\author[1,*]{Andres Potapczynski}
\author[1]{Andrew Gordon Wilson}
\affil[1]{%
    Center for Data Science\\
    New York University
}
\affil[*]{Equal contribution.}
  
\usepackage{xcolor}
\definecolor{dark-blue}{rgb}{0.15,0.15,0.4}
\definecolor{medium-blue}{rgb}{0,0,0.5}
\hypersetup{
   colorlinks, linkcolor={dark-blue},
   citecolor={dark-blue}, urlcolor={medium-blue}
}
  
\makeatletter
\newcommand*{\addFileDependency}[1]{
	\typeout{(#1)}
	\@addtofilelist{#1}
	\IfFileExists{#1}{}{\typeout{No file #1.}}
}
\makeatother

\begin{document}
\maketitle

\begin{abstract}
  Low-precision arithmetic has had a transformative effect on the training of neural networks, reducing computation, memory and energy requirements. However, despite its promise, low-precision arithmetic has received little attention for Gaussian processes (GPs), largely because GPs require sophisticated linear algebra routines that are unstable in low-precision. We study the different failure modes that can occur when training GPs in half precision. To circumvent these failure modes, we propose a multi-faceted approach involving conjugate gradients with re-orthogonalization, mixed precision, and preconditioning. Our approach significantly improves the numerical stability and practical performance of conjugate gradients in low- precision over a wide range of settings, enabling GPs to train on $1.8$ million data points in $10$ hours on a single GPU, without any sparse approximations.  
\end{abstract}

\section{Introduction}

Low-precision computations can profoundly improve training time, memory requirements, and energy consumption. Moreover, as hardware accelerators continue to develop, the scalability gains of low-precision algorithms will continue to appreciate into the future.
However, training in low-precision can increase the magnitude of round-off errors and decrease the supported range of the numerical representation, leading to lower quality gradients and ultimately sacrificing performance or, in the worst cases, convergence of the training procedure.

Recent works have developed efficient and stable methods for training deep neural networks in low-precision to great effect, often via the development of new floating point representations \citep{intel_bfloat16_2018,wang_bfloat16_2019}, mixed precision representations \citep{micikevicius2017mixed,das2018mixed,yang2019swalp}, or alternative summation techniques \citep{coorporation2017nvidia,micikevicius2017mixed,zamirai2021revisiting}.

Gaussian processes (GPs) are significantly more expensive than neural networks, and thus potentially stand more to gain from low-precision computations. GPs typically require $\mathcal{O}(N^3)$ computations and $\mathcal{O}(N^2)$ memory, for $N$ data points. 
However, GPs typically require sophisticated algebraic operations such as Cholesky decompositions for solving linear systems, which suffer badly from round-off error and are very poorly suited to half precision. Developing these approaches for low-precision computation is an open area of research \citep{higham2021exploiting}.

Instead, we focus on iterative techniques for Gaussian processes \citep{Gibbs97efficientimplementation,cutajar2016preconditioning,gardner2018gpytorch} that use conjugate gradients to compute the solutions to linear systems while exploiting extremely efficient MapReduce matrix vector multiplications (MVMs) using KeOps \citep{charlier2021kernel,feydy2020fast}. These approaches are particularly appealing in the context of low-precision computations: (1) they suffer less from round-off errors than Cholesky decompositions \citep{gardner2018gpytorch}; (2) they are easy to parallelize on GPUs, which are being designed specifically to accelerate low-precision computations \citep{gardner2018gpytorch, wang2019exactgps}; (3) they reduce memory consumption, which is a major \emph{computational} bottleneck, since lower memory consumption reduces the number of computationally expensive matrix partitions \citep{wang2019exactgps}.

While there are many challenges to overcome, we demonstrate that training GPs in half precision can be stable, efficient, and practical. In particular:
\begin{itemize}
    \item We numerically show that half precision kernel matrices have qualitatively different eigenspectra than kernels in higher precisions, and represent covariances at much shorter distances.
    \item We empirically investigate the effect of kernel choice, CG tolerance, preconditioner rank, and compact support on stability in half precision.
    \item Based on these insights, we propose a new numerically stable conjugate gradients (CG) algorithm that converges quickly in half precision, particularly due to the use of re-orthogonalization, mixed precision and the logsumexp trick.
    \item We provide a powerful practical implementation of Gaussian processes in half precision that reduces training times on $1.8$ million data point datasets to less than $10$ hours on a single GPU, a major advance over recent CG milestones of $1$ million data points in $3$ days on 8 GPUs \citep{wang2019exactgps}.
    \item We release code at \url{https://github.com/AndPotap/halfpres_gps}. 
\end{itemize}

\section{Preliminaries}

We briefly introduce background on different floating point precision representations, and on Gaussian processes. We also review how solving linear systems can be viewed as an optimization problem, and present conjugate gradients as an accelerated linear solver.

\subsection{Numerical Precision}
Modern computer architectures represent numbers in floating point precision, often with either $32$-bit (single) or $64$-bit (double) precision \citep{kahan1996ieee}.
\emph{Single precision} uses $1$ bit for the sign of the number ($s$), $8$ bits for the exponent ($e$), and $23$ bits for the mantissa (or significant, $m$).
Then, the value of a number is represented as 
\begin{equation*}
    \begin{split}
      \text{value}\left(s,e,m\right) = \left(-1\right)^{s} \times 2^{e-127} \times 1.m
    \end{split}
\end{equation*}
where $s \in \{0, 1\}$, $e \in \{0, \cdots, 255\}$ and $m \{1, \cdots, 1 + \sum_{i=1}^{23} 2^{-i}\}$. 
\emph{Single precision} can then represent values from $\pm \left(2 - 2^{-23}\right) \times 2^{127} \approx \pm 3.4028235 \times 10^{38}$ and measure small numbers, in absolute value, of the order of $2^{-126} \approx 1.175 \times 10^{-38}$. 
The round-off error is approximately of $2^{-24} \approx 6 \times 10^{-8}$ (Chapter 2.5, \citet{watkins2010}).

In contrast, \emph{half precision}, a representation with $16$-bits, uses $1$ bit for the sign, $5$ bits for the exponent and $10$ bits for the mantissa (or significant). Using again $s,e$ and $m$ to represent the previous bits, we obtain the following representation
\begin{equation*}
    \begin{split}
      \text{value}\left(s,e,m\right) = \left(-1\right)^{s} \times 2^{e - 14} \times 1.m
    \end{split}
\end{equation*}
where $s \in \{0, 1\}$, $e \in \{0, \cdots, 255\}$ and $m \{1, \cdots, 1 + \sum_{i=1}^{10} 2^{-i}\}$. 
Immediately we see that the range of values decreases to $\pm \left(2 - 2^{-10}\right) \times 2^{17} \approx \pm 2.620 \times 10^{5}$ and that the lowest value that can be represented is $2^{-14} \approx 6.1035 \times 10^{-5}$. 
The round-off error is now significantly higher as well, on the order of $2^{-14} \approx 10^{-4}.$
Much modern hardware attempts to reduce the round-off error via fused multiply-and-accumulate (FMAC) compute units to accumulate operations at higher precisions \citep{coorporation2017nvidia,zamirai2021revisiting,micikevicius2017mixed}.
We also note the existence of the bfloat16 standard \citep{intel_bfloat16_2018}, which uses a slightly different representation, preserving the range of single precision, while using fewer bits in the mantissa, thereby trading off the range for increased round-off error.

\subsection{Gaussian Processes}
We consider regression tasks on observed data $\mathcal{D} = \set{\left(\bm{x}_{i}, y_{i}\right)}_{i=1}^{N}$ for $\bm{x}_{i} \in \mathbb{R}^{D}$ and $y_{i} \in \mathbb{R}$. 
The Gaussian process (GP) model for the data is
\begin{equation*}
    \begin{split}
      f\left(\cdot\right) 
      &\sim 
      \mathcal{GP}\left(m\left(\cdot\right), k\left(\cdot, \cdot\right)\right),
      \\
      y_{i} 
      &=
      f\left(\bm{x}_{i}\right) + \epsilon_{i}, \quad \epsilon_{i} \sim \mathcal{N}\left(0, \sigma^{2}\right)
    \end{split}
\end{equation*}
where $k_\theta \left(\cdot, \cdot\right)$ is the covariance kernel, $m\left(\cdot\right)$ is the mean function (in the next equations we assume it set to zero without loss of generality) \citep{rasmussen_gaussian_2008}.
We train the kernel hyperparameters, $\theta,$ by minimizing the 
negative log marginal likelihood:
\begin{equation} \label{eq:loss}
    \begin{split}
      \\
      \mathcal{L}\left(\bm{\theta}\right)
      &= -\log p\left(\bm{y} | \bm{X} ;\bm{\theta}\right)
      \\
      &= \frac{1}{2} \log \left|\bm{\tilde K}\right| + \frac{1}{2} \bm{y}^{T} \bm{\tilde K}^{-1} \bm{y} + \frac{N}{2} \log\left(2 \pi\right)
    \end{split}
\end{equation}
where $\bm{\tilde K} \in \mathbb{R}^{N \times N}$ is the Gram matrix of all data points with diagonal observational noise
\begin{equation*}
    \begin{split}
      \bm{\tilde K}_{i,j} = (\bm K + \sigma^2 I)_{i,j} = k\left(\bm{x}_{i}, \bm{x}_{j}\right) + \sigma^{2} \mathbb{I}_{i = j},
    \end{split}
\end{equation*}
representing the evaluated covariance matrix as $\bm K.$
To perform hyperparameter estimation of $\bm{\theta}$, we need to compute the gradient of the loss function, which is given by 
\begin{equation} \label{eq:gradient}
    \begin{split}
      \nabla_{\bm{\theta}} \mathcal{L}
      = \frac{1}{2} &\tr\left(\bm{\tilde K}^{-1} \nabla_{\bm{\theta}} \bm{\tilde K}\right) 
      - \frac{1}{2} \bm{y}^{T} \bm{\tilde K}^{-1} \left(\nabla_{\bm{\theta}}\bm{\tilde K}\right) \bm{\tilde K}^{-1} \bm{y} \\
      \approx \frac{1}{M} \sum_{i=1}^M &\bm z_i^\top \bm{\tilde K}^{-1} \nabla_{\bm{\theta}} \bm{\tilde K} \bm z_i) 
      - \frac{1}{2} \bm{y}^{T} \bm{\tilde K}^{-1} \left(\nabla_{\bm{\theta}}\bm{\tilde K}\right) \bm{\tilde K}^{-1} \bm{y}
    \end{split}
\end{equation}
where the first line is the result of classical matrix derivatives \citep{rasmussen_gaussian_2008} and the second line comes from the application of Hutchinson's trace estimator \citep{Gibbs97efficientimplementation,gardner2018gpytorch}.
The limiting time complexity is that of computing $\bm v = \bm{\tilde K}^{-1} \bm y,$ which costs $\mathcal{O}(N^3)$ operations and space when using the Cholesky decomposition \citep{rasmussen_gaussian_2008}. 
Instead, \citet{Gibbs97efficientimplementation} and then
\citet{gardner2018gpytorch},
proposed to use conjugate gradients (CG) to compute these linear systems, reducing the time complexity down to $\mathcal{O}(JN^2),$ for $J$ steps of conjugate gradients.

\subsection{Conjugate Gradients}

We use conjugate gradients (CG) to solve the linear system $\bm A \bm x = \bm b$ for $\bm x$, so that $\bm x = \bm A^{-1} \bm b$ \citep{hestenes1952methods,nocedal2006,demmel1997applied}.
CG uses a three-term recurrence where each new term requires only a matrix-vector multiplication with $\bm{A}$. Formally, each CG iteration computes a new term of the following summation:
$\bm{A}^{-1}\bm{b} = \sum_{i=1}^{N} \alpha_{i} \bm{d}_{i},$
where, for simplicity, we assume that the algorithm is initialized at $\bm{x}_{0} = \bm{0}$, $\alpha_{i}$ are the step-size coefficients and $\bm{d}_{i}$ are the orthogonal conjugate search directions \citep{golub2018matrix}. In infinite precision representation, $N$ iterations of CG produce all the $N$ summation terms and recover the exact solution. 
The CG algorithm is shown in Algorithm \ref{alg:CG_basic} in the Appendix.

\section{Related Work}
\textbf{Iterative Gaussian Processes} \quad
There has been extensive research on alternative algorithms to reduce the cubic runtime complexity of training Gaussian processes via Cholesky.
In this work, we primarily focus on iterative GPs methods, prioritizing over other alternative approaches.
\citet{Gibbs97efficientimplementation} studied iterative techniques for GPs by using conjugate gradients to solve the linear systems that result from training GPs and also to estimate the stochastic trace term that appears when computing the loss. 
In contrast to Cholesky, these iterative approaches reduce the overall complexity of GP regression to $\mathcal{O}(JN^2)$ for $J$ steps of CG. \citet{cutajar2016preconditioning} re-visited these approaches, and applied preconditioners based on low-rank kernel approximations to increase the convergence speed to the linear system solutions, using double precision.
\citet{gardner2018gpytorch} proposed the batch conjugate gradients algorithm that we extend in this paper and used Lanczos to estimate log determinants, focusing their efforts in single precision. 
\citet{wang2019exactgps} extended this work and enabled exact GPs to be trained on $1$ million data points by using $8$ GPUs over $3$ days via data partitioning.
\citet{meanti2020kernel} also used KeOps over several GPUs for Nystr\"{o}m-based kernel regression, achieving results on datasets of up to $1$ billion data points. 

\textbf{Lower Precision Arithmetic} \quad 
Interest in training neural network models in lower precisions than the traditional double or single precisions has been around as early as \citet{hammerstrom1990vlsi}.
In the modern era, \citet{gupta2015deep} and \citet{chen2014dadiannao} pioneered the usage of lower precision arithmetic 
to reduce memory costs so that larger deep neural network models could be trained.
\citet{gupta2015deep} proposed the usage of stochastic rounding to reduce errors from the loss of precision when moving to half (e.g. 16-bit arithmetic) precision down from single precision, while \citet{chen2014dadiannao} used special representations to enable accurate training of DNNs.
These works have led to a flurry of research into \emph{mixed precision} training of deep neural networks \citep{micikevicius2017mixed,das2018mixed,yang2019swalp}.
In mixed precision training, the activations and gradients of each DNN layer are propagated in half but the weights are stored in single precision \citep{micikevicius2017mixed}.
The success of these lower precision approaches has led to significant efforts in reducing the memory overhead even further via quantization of neural network layers \citep{jacob2018quantization,das2018mixed}, the development of specialized chip architectures that speed up lower precision arithmetic such as modern GPUs \citep{coorporation2017nvidia} and TPU cores \citep{wang_bfloat16_2019},
and new standards that enhance mixed precision training such as bfloat16 \citep{wang_bfloat16_2019,intel_bfloat16_2018}.
The use of lower precision arithmetic has found some applications outside of deep learning for kernel approximations. 
For example, using quantized random Fourier features as in \citep{pmlr-v89-zhang19f,li2021quantization} or inside of structured basis functions like Fastfood \citep{le2013fastfood,yang2015carte}.

\textbf{Improving CG convergence} \quad
When using infinite precision arithmetic, CG is guaranteed to converge relatively fast to the solution of the linear system. 
However, the round-off error introduced when using finite precision affects the convergence to the solution.
There are several strategies that can generally improve the stability and convergence of CG, such as: preconditioning, re-orthogonalization, mixed precision and blocked arithmetic.
For double precision GP inference, 
\citet{cutajar2016preconditioning} found that preconditioning was necessary for the CG solves to be accurate,
a finding echoed by \citet{gardner2018gpytorch} and more recently \citet{wenger2021reducing}.
\citet{gardner2018gpytorch},
\citet{wang2019exactgps} and 
\citet{wenger2021reducing} 
proposed using pivoted Cholesky preconditioners \citep{harbrecht2012lowrank}.
Additionally, recent work in numerical methods, such as \citet{gratton2019minimizing}, has argued for the use of re-orthogonalization in CG as a general numerical strategy. 
\citet{haidar2017investigating} and \citet{haidar2018harnessing} alternatively proposed iterative refinement inside of GMRES (a method related directly to CG) for solving dense linear systems on GPUs in lower precision arithmetic, while \citet{abdelfattah2020investigating} considered mixed precision solves using iterative refinement.
\citet{higham2021numerical} and \citet{higham2021exploiting} argue for blocked arithmetic to reduce round-off errors, which we demonstrate has considerable advantages in Figure \ref{fig:mm_accuracy}. 

\section{What happens to kernel matrices in half precision?}\label{sec:understanding}
\begin{figure}[!t]
\centering
    \centering
    \includegraphics[width=\linewidth]{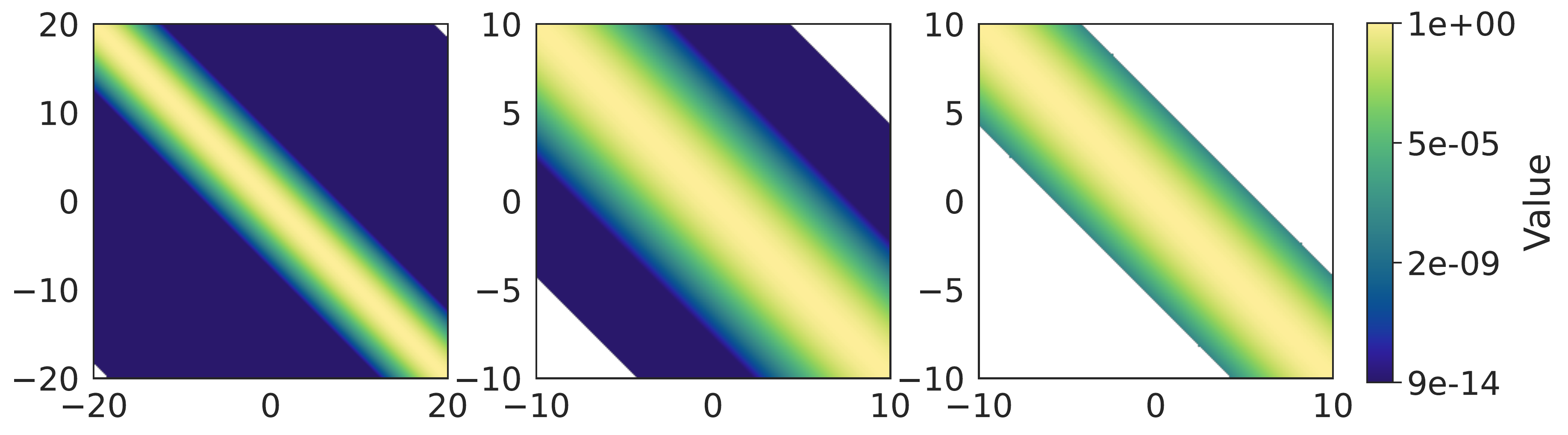}
        \caption{Elementwise logarithms of RBF kernel matrices in double (left), single (middle), and half (right) precisions. As we move to lower precision, the support of the kernel matrices becomes increasingly truncated, which can be exploited for faster inference. White values show where the kernel matrix is exactly zero, producing a kernel with compact support. A compactly supported kernel produces more efficient kernel matrix MVMs as we can safely remove data points that have distances from a given test point larger than can be represented in a given numerical precision.
        }
    \label{fig:rbf_kernels}
\end{figure}

 In this section, we investigate the properties of half precision kernels, finding that they have a truncated support, that their eigenspectra are qualitatively different than in single precision (leading to different generalization properties) and that direct methods such as the standard Cholesky decomposition fails in half precision due to round-off error. In Section~\ref{sec: methods} we leverage these insights to develop effective methods for low-precision GP inference. 

\subsection{Finite precision kernels have finite support}\label{sec:support}
First, in Figure \ref{fig:rbf_kernels}, we display the elementwise logarithms of RBF kernel matrices of size $500$ as we lower the precision from double (left) to single (middle) to half (right) precision. 
The elementwise logarithm allows us to see that there are significantly more values of the kernel matrix represented as \emph{exactly zero}, displayed as white, for the half precision kernels than for the higher precisions. 
For the half and single precision kernels, we consider data evenly spaced in $[-10,10]$ while for double we consider data in $[-20, 20].$

\begin{figure*}[!t]
\centering
\begin{subfigure}{0.19\textwidth}
    \centering
    \includegraphics[height=2.5cm]{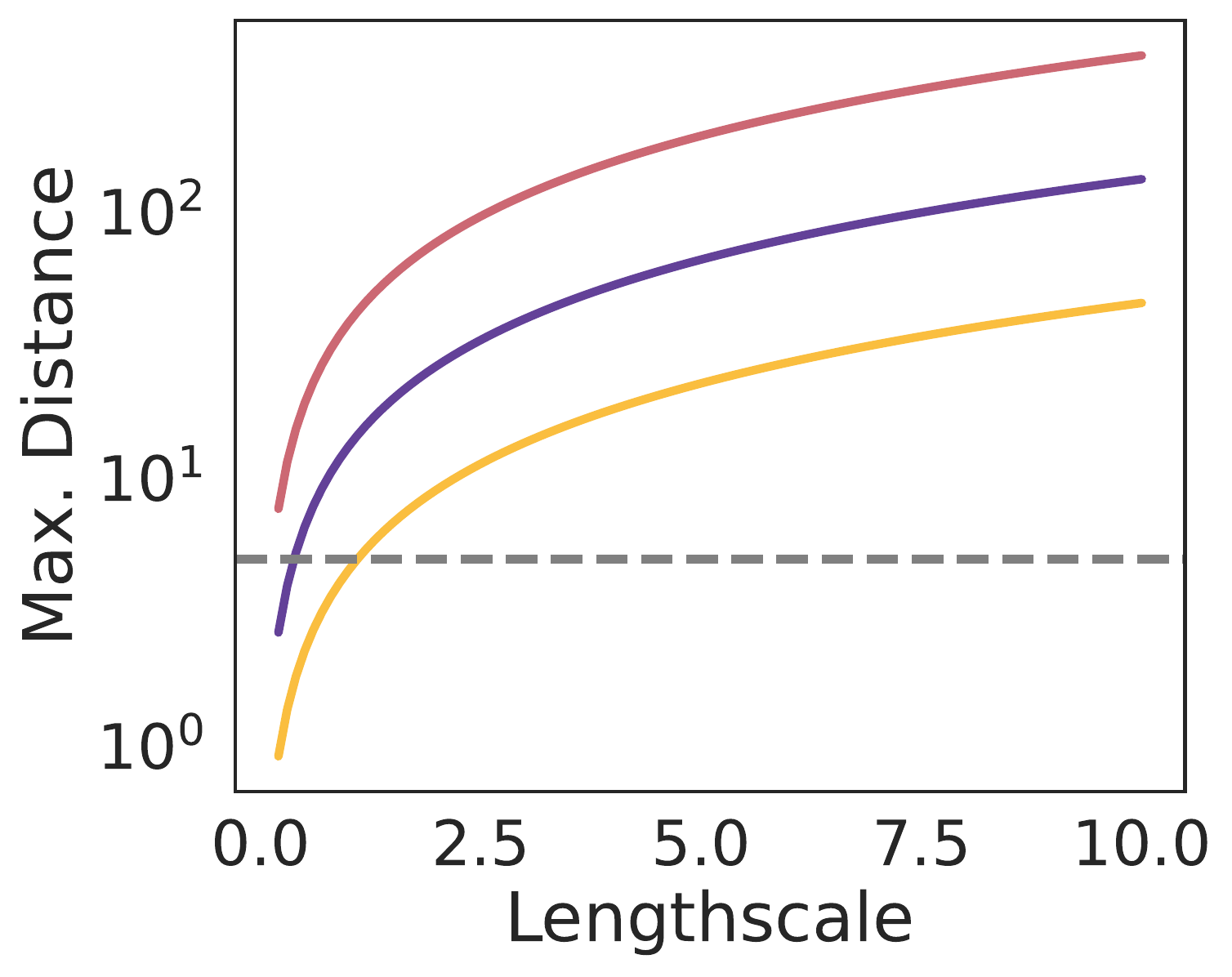}
    \caption{RBF.}
    \label{fig:max_distances_rbf}
\end{subfigure}
\begin{subfigure}{0.19\textwidth}
    \centering
    \includegraphics[height=2.5cm]{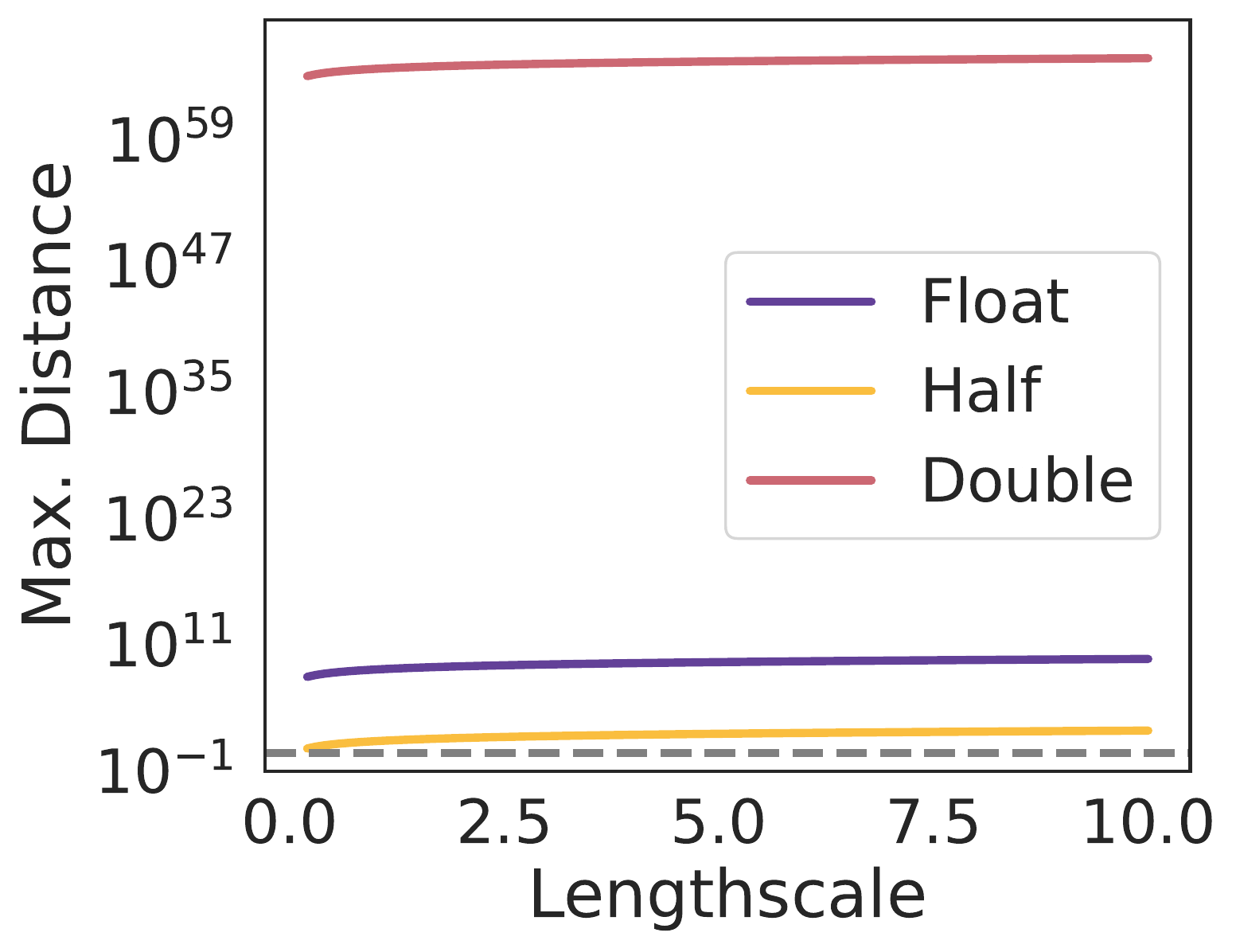}
    \caption{Rational quadratic.}
    \label{fig:max_distances_rq}
\end{subfigure}
\begin{subfigure}{0.19\textwidth}
    \centering
    \includegraphics[height=2.5cm]{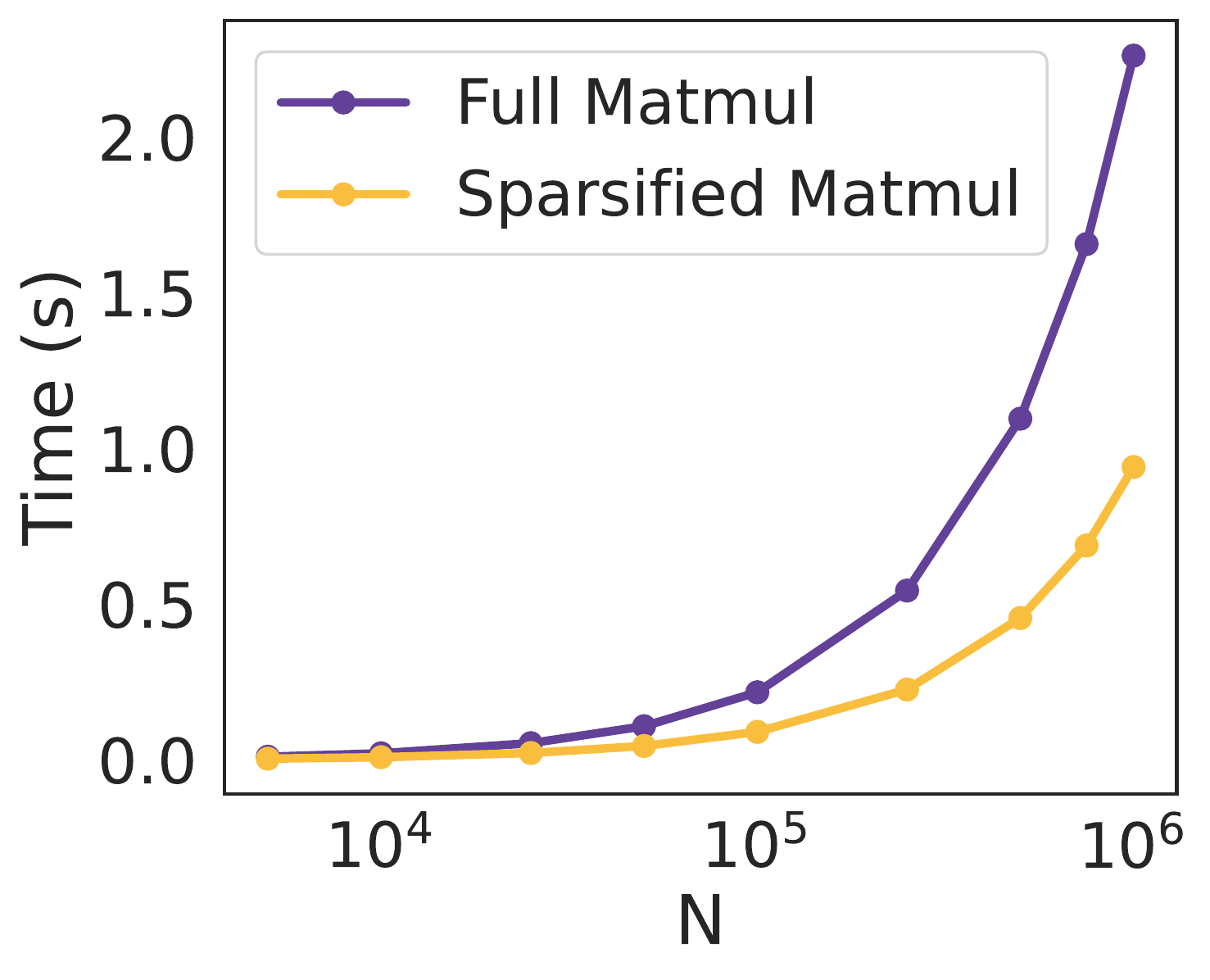}
    \caption{Time of MVMs.}
    \label{fig:sparsity_exp}
\end{subfigure}
\begin{subfigure}{0.19\textwidth}
    \centering
    \includegraphics[height=2.6cm]{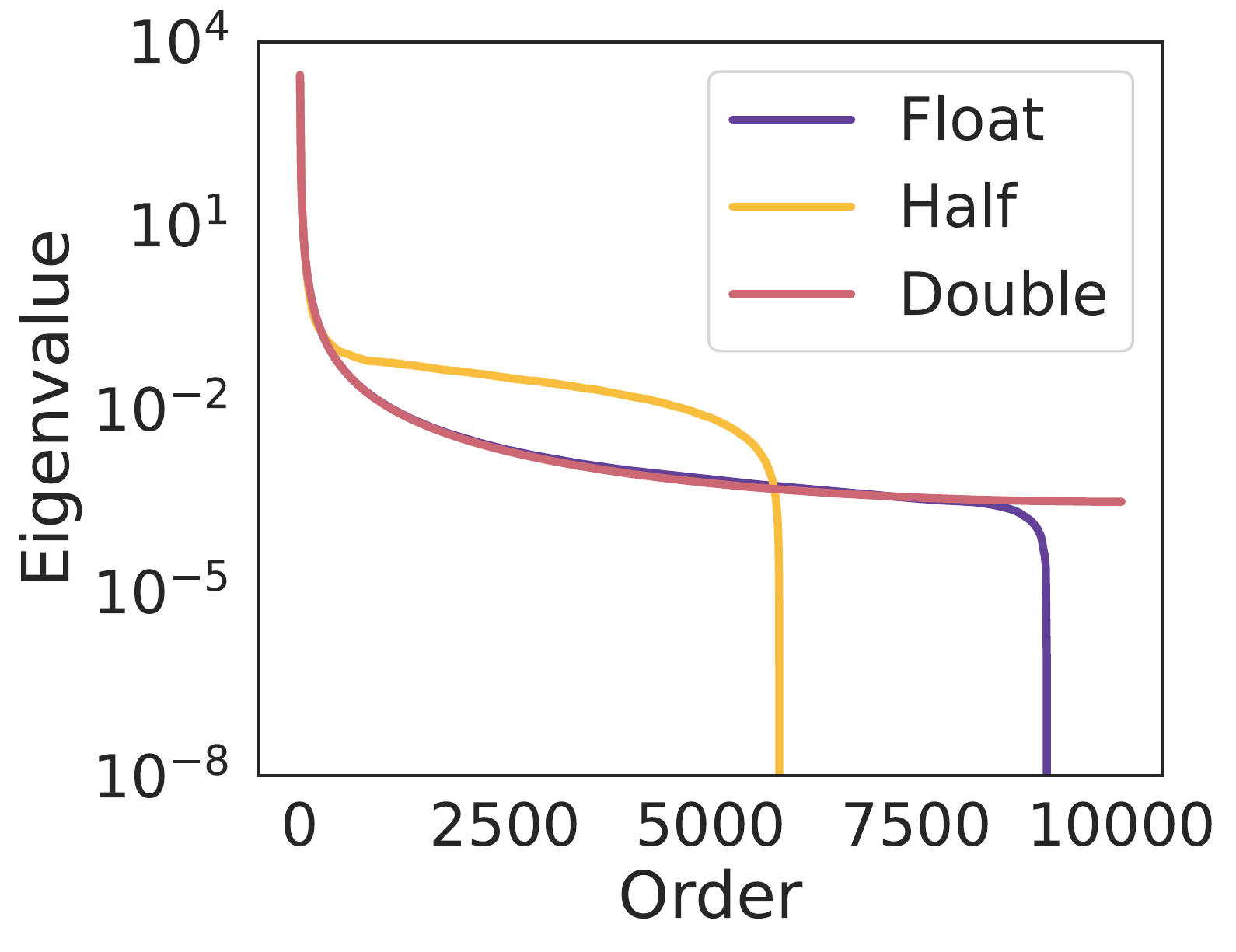}
    \caption{Eigenvalue spectrum.}
    \label{fig:matern_spectrum}
\end{subfigure}
\begin{subfigure}{0.19\textwidth}
    \centering
    \includegraphics[width=\textwidth]{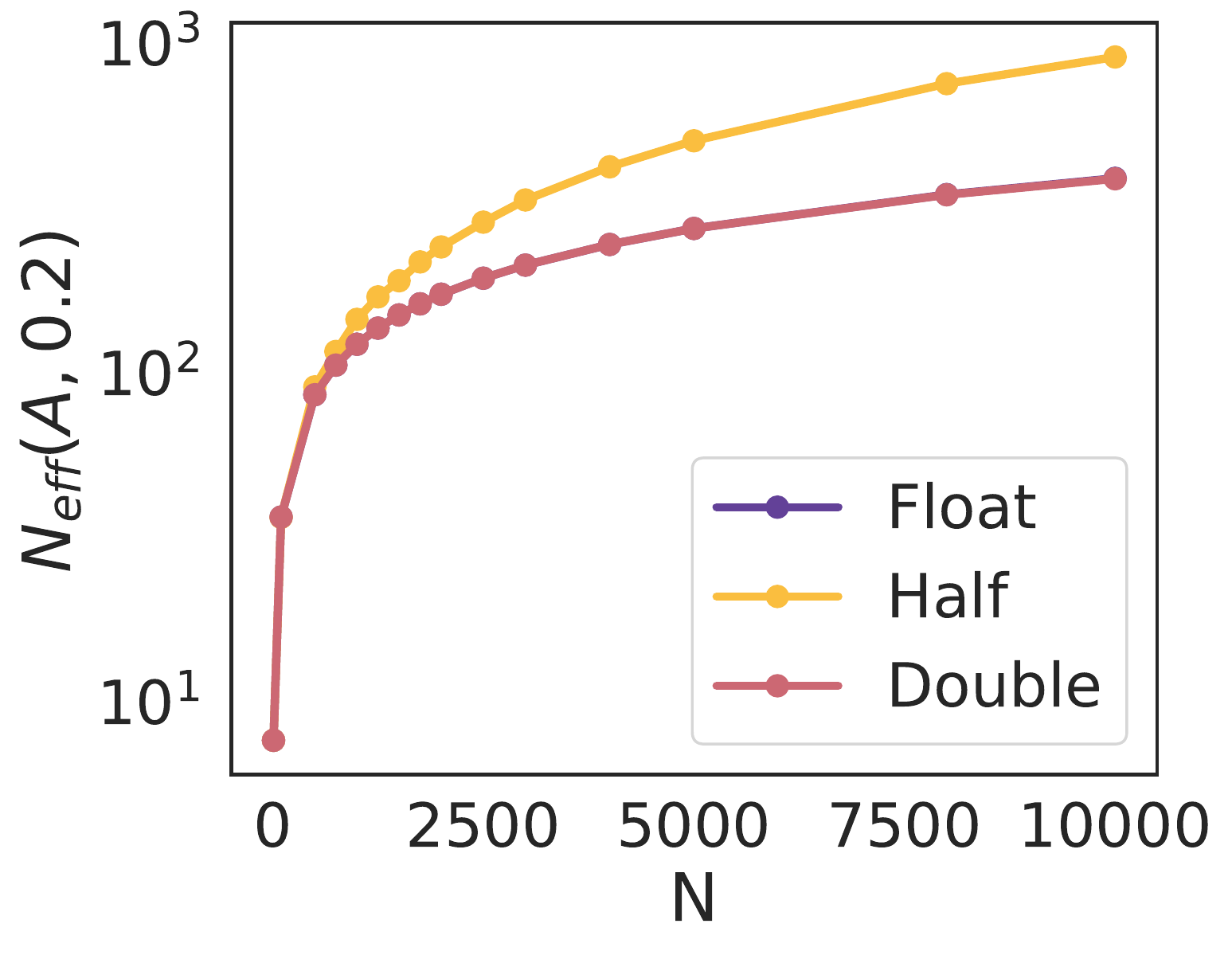}
    \caption{Effective dimension.}
    \label{fig:matern_ed}
\end{subfigure}
\caption{\textbf{(a, b)} Maximum representable distance across precisions for RBF and rational quadratic (RQ, $\alpha=5$) kernels. Shown as a gray line for reference is $5$ units of distance. RBF kernels have significantly shorter representable distances than RQ kernels. \textbf{(c)} Time for both the full and truncated MVMs as the dataset size grows. The truncated MVM grows at a significantly lower rate than the full MVM because it only operates on one quarter of the full matrix. \textbf{(d)} An observed eigenvalue spectrum for a Mat\'ern-$1/2$ kernel across precisions; the kernel evaluated in half precision has a much slower rate of decay in its spectrum than either single or double, which produces a much larger effective dimensionality \textbf{(e)} across matrix sizes.
}
\end{figure*}

We can exactly quantify the range of support of common stationary kernel matrices for a given lengthscale. 
For example, for a Mat\'ern-$1/2$ kernel with lengthscale, $l$, the kernel function will be exactly zero when $k(x, x') = k(d) = \exp\{-d / l\} < \epsilon$ (satisfied when $d > -\log \epsilon / l$), where $\epsilon$ is the smallest representable value in that given precision. We provide analogous results for other kernels in the Appendix.
If we have $d = 0.75$ (equivalent to $x = -0.5$ and $x' = 0.25$), then we cannot represent a relationship in half precision between the two points for lengthscales greater than about $9.24 \approx -\log \epsilon / 0.75.$
We show the maximum distance representable in double, single and half precisions for RBF (Figure \ref{fig:max_distances_rbf}) and rational quadratic kernels (Figure \ref{fig:max_distances_rq}, displaying $5$ units (a proxy for the maximum distance of input data standardized to have mean $0$ and variance $1$) as a gray dashed line. 
RBF kernels have the smallest support, especially for short lengthscales, followed by Mat\'ern kernels (see Figure \ref{fig:max_distances_matern}), 
and then rational quadratic kernels.

We can further understand these results by examining the spectrum of the eigenvalues of RBF kernels. 
In infinite precision, an RBF kernel has support over the full space, with $\lambda_k > 0$ for each $k$. 
For $k \geq \mathcal{O}(\log \delta)$
with $\delta$ representing the round-off error in the specified precision, then $\lambda_k = 0$ in finite precision, empirically reducing the support of the kernel (see Appendix \ref{app:theory} for further details).
These results are reminiscent of how probabilities of a Gaussian variable taking values several standard deviations from its mean are numerically zero, due to the sharp decay of the tails of the distribution.

Kernels represented with compact support can be exploited for scalable computations.
Indeed, we demonstrate the potential for this type of structure exploitation when computing GP predictive means in Figure \ref{fig:sparsity_exp}; first, we compute the predictive mean cache $\bm{v} = \bm{\tilde K^{-1}} \bm y$ before either computing a full predictive mean $\mu_{\text{full}} = K(x^*, \bm X) \bm v$ or a truncated predictive mean by dropping all data points that will be represented as zero in half precision, e.g. $\mu_{\text{trunc.}} = K(x^*, \bm X_{\text{close}}) \bm v_{\text{close}},$ which drops approximately $3/4$ of the data in this example.
As shown in Figure \ref{fig:sparsity_error}, the error is zero, indicating the predictive mean does not change at all!

\subsection{Investigating the eigenspectrum}\label{sec:eigenspectrum}

We next investigate the eigenvalue spectrum across precisions, considering a Mat\'ern-$1/2$ kernel, and data drawn from $[-3,3]$, displaying the empirical spectrum in Figure \ref{fig:matern_spectrum}.
First, we see that the condition number of the linear system is about the same across precisions, since the maximum eigenvalue is the same (see Figure \ref{fig:other_spectra}), and we know that for GP kernel matrices the smallest eigenvalue is really close to the noise hyperparameter value $\sigma^2$.
Second, we see that the eigenvalues of the half precision kernel decay slower before dropping to zero when compared to other precisions.
As the per iteration progress in CG is bounded 
by the difference between the current iteration eigenvalue and the first (Thm 5.5 of \citet{nocedal2006}), then we would
expect CG to take more iterations to converge in half precision as the progress of each step in CG is minimal if the eigenvalues are similar \citep{demmel1997applied}.

Furthermore, the difference between the eigenvalue spectrum of single and half precision may lead to worse generalization.
This point can be argued through the effective dimension of a kernel matrix $\bm K$ defined as 
$N_{\text{eff}}(\bm{K}, \sigma^2):=\sum_{i=1}^N \frac{\lambda_i}{\lambda_i + \sigma^2}$ where $\lambda_i$ represents the eigenvalues of $\bm K$.
First, note that the effective dimension is a critical term in generalization bounds of kernel regression and GP methods  \citep[e.g.,][]{zhang2005learning,opper1998general}. These generalization bounds are bounded above by $N_{\text{eff}}(\mathbf{K}, \lambda)/N$;
therefore, higher effective dimensionality leads to looser bounds. We prove in Appendix \ref{app:theory} that 
\begin{align}\label{eq:exp}
    \mathbb{E}\left(\sum_{i=1}^N \frac{Q(\lambda_i)}{Q(\lambda_i) + s}\right) \geq N_{\text{eff}}(\mathbf{K}, s),
\end{align}
where $Q(\lambda_i)$ denotes the quantization error of $\lambda_i$ and, following \citet{li2019dimension}, we assume that
$Q(\lambda_i)$ is distributed uniformly $U(\lambda_i - \delta, \lambda_i + \delta)$, where $\delta$ represents the round-off error of
our numerical precision. We then argue in Appendix \ref{app:theory} how the LHS of equation \ref{eq:exp} increases for lower precisions (that is, for higher $\delta$) leading to provably worse generalization.
Second, it has been empirically shown how for similar training losses, lower effective dimensionality correlates with better generalization
\citep{zhang2005learning,caponnetto_optimal_2007, maddox2020rethinking}. In Figure \ref{fig:matern_ed} we empirically observe that the slower eigenvalue decay
of half precision implies that, as $N$ increases, the effective dimensionality increases much faster for the half precision kernel.
This finding is consistent with our previous theoretical analysis and also suggests that half precision will have a worse generalization than single precision in the context of kernel methods.

\begin{figure}[!t]
\centering
\includegraphics[width=0.4\linewidth]{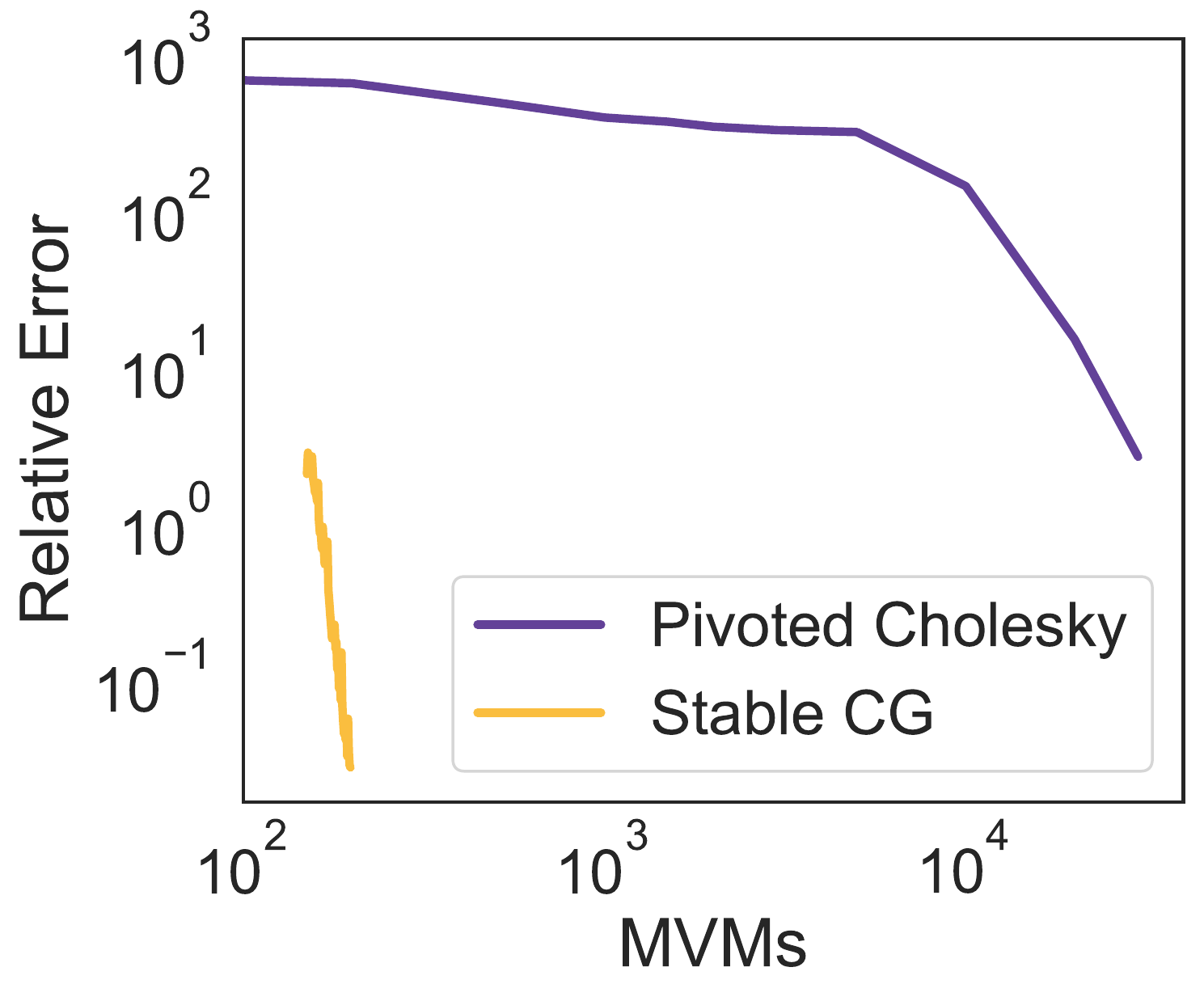}
\caption{
Pivoted Cholesky fails to solve a linear system in half, whereas 
preconditioned CG solves the system efficiently and accurately. 
}
\label{fig:protein_pc_conv}
\end{figure}

\subsection{What about Direct Methods?}\label{sec:pc_exp}
Can direct methods, such as Cholesky decompositions, effectively solve linear systems in low-precision? Support for Cholesky factorizations in half precision is not included in GPU based linear algebra libraries (LAPACK) and approaches for half precision Cholesky factorizations use iterative techniques on the backend \citep{higham2021exploiting}.
We studied the effectiveness of direct methods
by using Eq. \ref{eq:woodbury} to solve linear systems on the UCI \emph{protein} dataset in half precision, varying the size of the preconditioner directly.
We find that the computations in Eq. \ref{eq:woodbury} accumulate a large number of run-off errors which results in high
residual norms of the resulting solves as seen in Figure \ref{fig:protein_pc_conv}.

\section{Our Method}
\label{sec: methods}

Our methodology expands upon the ideas of \citet{gardner2018gpytorch} and
\citet{wang2019exactgps}
for training GPs solely through matrix-vector operations via CG. We modify the CG algorithm in order to support half precision by increasing the numerical stability through:
scale translation, mixed precision aggregation and re-orthogonalizaton.

\subsection{Half Precision Matrix Vector Multiplies}

Following \citet{wang2019exactgps}
and \citet{meanti2020kernel}, 
we exploit KeOps \citep{charlier2021kernel} to produce our \emph{matrix-free} approach.
Where \emph{matrix-free} refers to the numerical strategy of evaluating matrix multiplies by generating, on the fly, the entries required by the operation
without the need to hold the matrix in memory.
For a more detailed explanation of KeOps, please see \citet{charlier2021kernel} and \citet{feydy2020fast}.
More specifically, to use conjugate gradients effectively, we need to evaluate $\bm{\tilde K} \bm{v}$ which is written as 
\begin{align}
    \begin{split}
      \bm{\tilde Kv}_{i} =
        a^{2} \sum_{j=1}^{N}  k\left(x_{i}, x_{j}\right) v_{j} + \sigma^{2} v_{i}.
    \end{split}
    \label{eq:mvm}
\end{align}
KeOps performs the inner loop of this multiplication via partitioning the resulting rows and columns of the matrix into blocks $B_1, \cdots B_k$ before reducing the resulting matrices, computing the row-wise operations in parallel. 
To exploit parallelism on accelerated hardware, we decompose the data into blocks $B_{1}, \dots, B_{K}$ of size $\left|B_{k}\right| = M << N$ ($M$ is the CUDA block size of $192$) such that
$\bigcup_{k=1}^{K} B_{k} = \set{1, \dots, N}$. 
Moreover, we decompose $\bm{Kv}$ into $K$ separate products $\bm{K}^{\left(k\right)} \bm{v}^{\left(k\right)}$ where the $\bm{K}^{\left(k\right)}$ is an
$M \times M$ matrix composed of 
$\bm{K}^{\left(k\right)}_{i,j} = a^{2} k\left(x_{i}, x_{j}\right) + \sigma^{2} \delta_{i,j}$ for all $i,j \in B_{k}$ and where $\bm{v}^{\left(k\right)}_{i} = \bm{v}_{i}$ for $i \in B_{k}$. We compute each separate product $\bm{K}^{\left(k\right)} \bm{v}^{\left(k\right)}$ as a regular matrix vector product (MVM). 
We note that a \emph{matrix-free} 
approach does involve the creation of block matrices but not of the full matrix $\bm{K}$ where this approach
only ever requires the storage of $M \times M$ block matrices at once.
For accurate matrix products, we use block summation in a higher precision as is common on GPUs \citep{zamirai2021revisiting,coorporation2017nvidia}.

\begin{figure}[!t]
    \centering
    \begin{subfigure}{0.22\textwidth}
    \includegraphics[width=\linewidth]{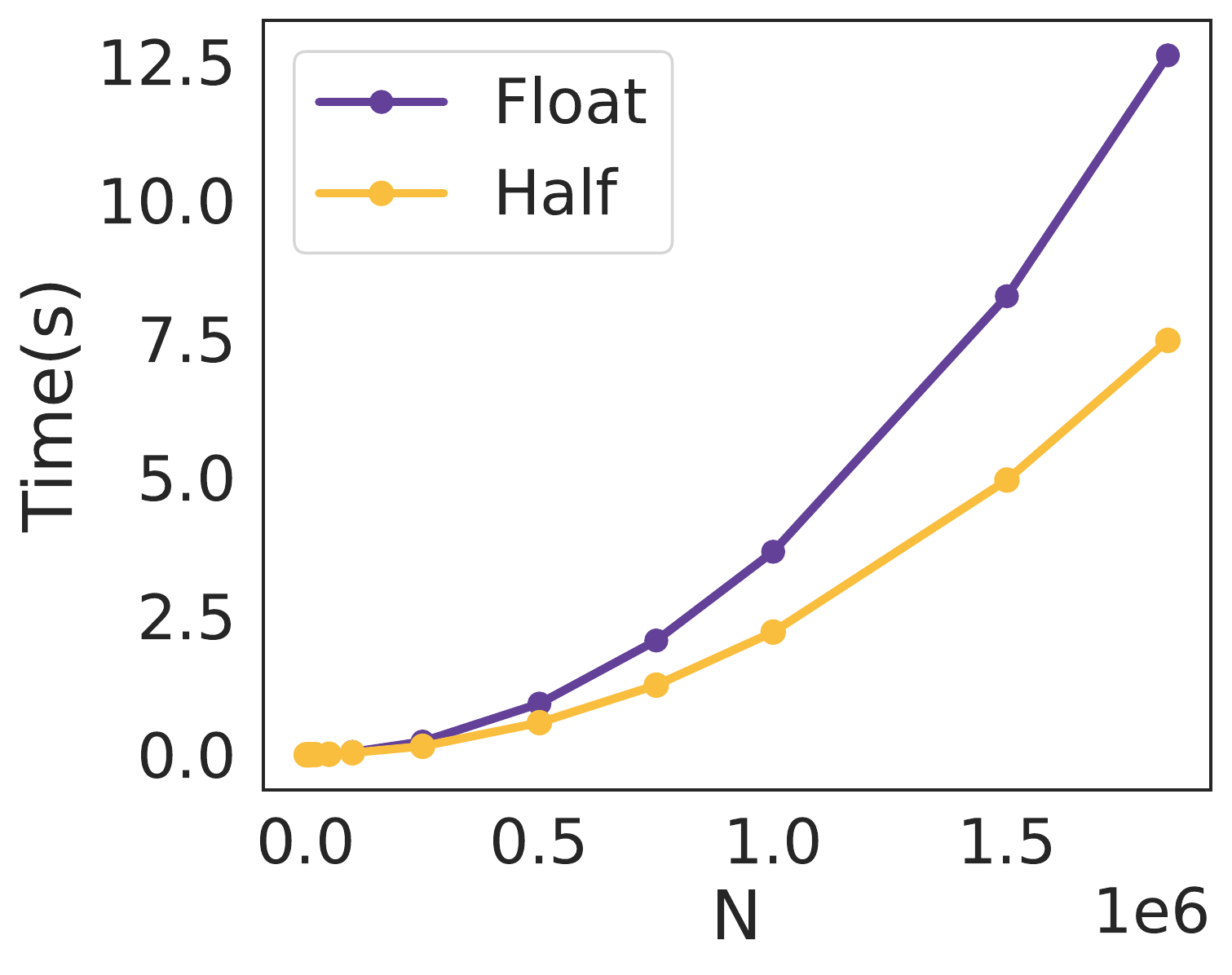}
    \caption{Time.}
    \label{fig:mm_times}
    \end{subfigure}
    \begin{subfigure}{0.22\textwidth}
    \includegraphics[width=\linewidth]{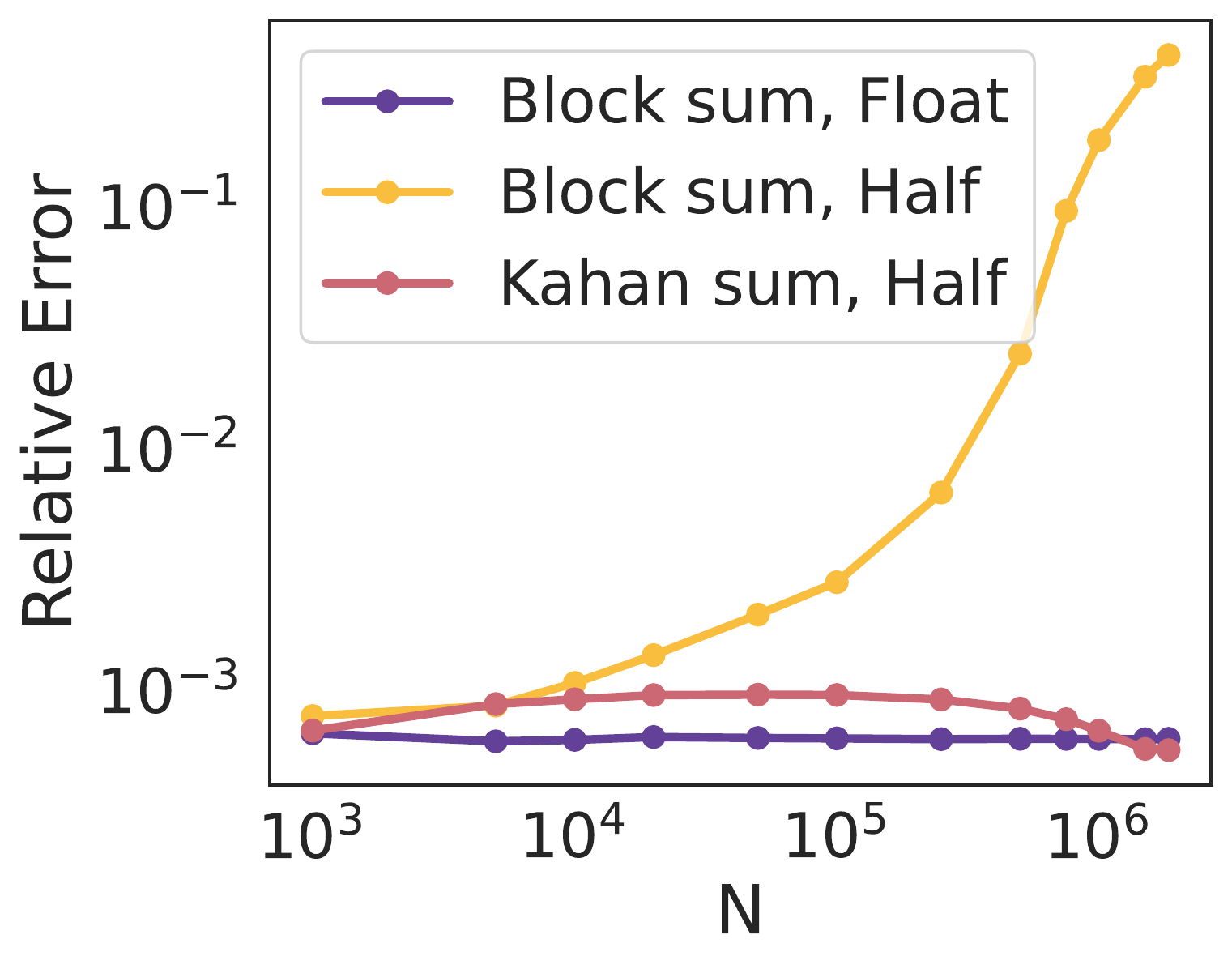}
    \caption{Accuracy.}
    \label{fig:mm_accuracy}
    \end{subfigure}
    \caption{\textbf{(a)} Timings for a single precision matrix vector multiplication on the HouseElectric dataset. Half is consistently about twice as fast as single precision. \textbf{(b)} Accuracies compared to the single precision for various schemes. Block summation performs each block within half precision, before doing accumulations in half or single and then casting the result back to half, while Kahan summation performs all operations in half. Block summation with single and Kahan summation are nearly as accurate as an entirely single precision MVM.
    }
\end{figure}

To further exploit parallelism and to reduce memory overhead, we compute all matrix vector products in half precision, rather than single point precision \citep{gardner2018gpytorch} or double precision.
Moving to half precision immediately produces about a $2$x speedup compared to half precision, as shown in Figure \ref{fig:mm_times}, as evaluated on dataset sizes successively larger on the $1.8$ million data point HouseElectric dataset from the UCI repository \citep{Dua:2019}.
Furthermore, as shown in Figure \ref{fig:mm_accuracy}, these matrix vector multiplications are highly accurate with relative errors averaging less than $0.1$\% when we perform block summation in single precision.
By comparison, block summation in half precision is significantly less accurate, while Kahan summation \citep{kahan65further} in half is nearly as accurate as block summation in single precision.
All three approaches have similar speeds as shown in the Appendix. 
Thus, by moving to half precision we have immediately achieved a nearly $2$x speedup in time complexity (and a similar reduction in memory) at negligible accuracy costs.

However, the limitations of the range can cause numerical overflow for large matrix vector multiplications. 
To see why this is problematic, we note that the largest value representable in half precision is $2^{13},$ while the output of a kernel matrix vector multiply scales quasi-linearly with the size of the kernel matrix, $N,$
\begin{equation*}
    \begin{split}
      \bm{\tilde K}\big(\bm{v}\big)_{i} 
      &=
      a^{2} \sum_{j=1}^{N}  k\left(x_{i}, x_{j}\right) v_{j} + \sigma^{2} v_{i}.
      \\
      &\leq
      N \left(a^{2} + \sigma^{2}\right) \norm{v}_{\infty}.
    \end{split}
\end{equation*}
We instead downscale every MVM by $N^{-1/2}$, computing $\bm{\tilde K}(\bm v / N^{1/2})$, which produces an upper bound that scales as $N^{1/2}$.

\begin{algorithm}[!t]
\caption{Enhanced Stability CG (\textcolor{blue}{blue font denotes differences from standard CG}) 
}
\label{alg:rCG}
\begin{algorithmic}[1]
  \STATE \textbf{Input:} MVM function $\bm{K}\left(\cdot\right)$, initial solution guess $\bm{x}_{0}$, linear system right hand side $\bm{b}$, 
  tolerance $\epsilon$, preconditioner function $\bm{P}\left(\cdot\right)$
  \STATE \textbf{Initialize:} $k \leftarrow 0$, $\bm{r}_{0} \leftarrow \bm{K}\left(x_{0}\right) - \bm{b}$, $\bm{d}_{0} \leftarrow -\bm{r}_{0}$,
  $\bm{z}_{0} = \bm{P}\left(\bm{r}_{0}\right)$
  and 
  \textcolor{blue}{$\log \gamma_{0} \leftarrow \text{L}\Sigma\text{E}\left(\bm{r}_{0}^{T} \bm{z}_{0}\right)$}.
  \WHILE{$\norm{\bm{r}_{k}}_{2} < \epsilon$}
    \STATE \textcolor{blue}{$\alpha_{k} = \exp\left(\log \gamma_{k} - \text{L}\Sigma\text{E}\left(\bm{d}_{k}^{T} \bm{K}\left(\bm{d}_{k}\right)\right) \right)$}
    \STATE $\bm{x}_{k+1} = \bm{x}_{k} + \alpha_{k} \bm{d}_{k}$
    \STATE $\bm{r}_{k+1} = \bm{r}_{k} + \alpha_{k} \bm{K}\left(\bm{d}_{k}\right)$
    \FOR{$j=0$ to $k$}
      \STATE \textcolor{blue}{$\bm{r}_{k+1} = \bm{r}_{k+1} -\left(\bm{u}_{j}^{T} \bm{r}_{k+1}\right) \bm{u}_{j}$}
    \ENDFOR
    \STATE $\bm{z}_{k+1} = \bm{P}\left(\bm{r}_{k+1}\right)$
    \STATE \textcolor{blue}{$\log \gamma_{k+1} = \text{L}\Sigma\text{E}\left(\bm{r}_{k+1}^{T} \bm{z}_{k+1}\right)$}
    \STATE \textcolor{blue}{$\beta_{k+1} = \exp\left(\log \gamma_{k+1} - \log \gamma_{k}\right)$}
    \STATE $\bm{d}_{k+1} = -\bm{r}_{k+1} + \beta_{k+1} \bm{d}_{k}$
  \ENDWHILE
\end{algorithmic}
\end{algorithm}

\subsection{Half Precision Conjugate Gradients}
\paragraph{Rescaling Conjugate Gradients}
The step sizes of the CG algorithm have a natural interpretation: $\alpha_{k}$ ensures we are minimizing the objective function along the path
$\bm{x}_{k} + \alpha_{k} \bm{p}_{k}$ and $\beta_{k}$ guarantees conjugacy between the search directions. The accelerated convergence of CG depends on the correct computation of $\alpha_{k}$ and $\beta_{k}$; however, in finite arithmetic we cannot compute these quantities exactly. Worse, the round-off error is amplified when using half precision as we compute the step size $\alpha_{k}$ and $\beta_{k}$ terms
\begin{equation*}
    \begin{split}
        \alpha_{k} 
        = \frac{\bm{z}_{k}^{T} \bm{r}_{k}}{\bm{d}_{k}^{T} \bm{K} \bm{d}_{k}}
        \quad \text{and } \quad
        \beta_{k+1}
        = \frac{\bm{z}_{k+1}^{T} \bm{r}_{k+1}}{\bm{z}_{k}^{T} \bm{r}_{k}}.
    \end{split}
\end{equation*}
We need to prevent round-off and overflow error with these terms, so we store them solely in the log-scale (as they are positive by definition).
To do so, we exploit the well known logsumexp trick, which states that 
for a given vector $\bm{w}$ and $\bm{z}$ such that $\bm{w}^{T} \bm{z} > 0$, to compute $\log \bm{w}^{T} \bm{z}$ we use the following transformation 
$\text{L}\Sigma\text{E}\left(\cdot\right)$ such that
\begin{equation*}
    \begin{split}
      \text{L}\Sigma\text{E}\left(\bm{w}^{T} \bm{z}\right)
      =
      y_{\text{max}} + \log \left(\sum_{i=1}^{N} s_{i} \exp \left(y_{i} - y_{\text{max}}\right)\right)
    \end{split}
\end{equation*}
where $y_{i} = \log \left|w_{i}\right| + \log \left|z_{i}\right|$ and $s_{i} = \text{sign}\left(w_{i} z_{i}\right)$.
For example, we compute and store $\log \alpha_k = \text{L}\Sigma\text{E}(\bm r_k^\top \bm z_k) - \text{L}\Sigma\text{E}(\bm d_k^\top \bm K \bm d_k).$

\paragraph{Re-orthogonalization}
In infinite precision, each direction vector $\bm{d}_{j}$ for CG is $\bm{K}$ orthogonal, that is $\bm{d}_{i}^{T} K \bm{d}_{j} = 0$ whenever $i \neq j$, 
producing residual vectors, $\bm r_k$, that are $\bm K$ orthogonal to each other \citep{nocedal2006}. In practice, orthogonality is lost due to round-off error leading to slower convergence or even divergences in the residual vectors \citep{gratton2019minimizing,cutajar2016preconditioning}. 
To accelerate convergence and preserve orthogonality, we apply explicit Gram-Schmidt re-orthogonalization inside CG \citep{gratton2019minimizing}.
Specifically, we re-orthogonalize the residual vector for time step $k$, updating the new residual $\bm{r}_{k}= \bm{K}\bm{x}_{k} - \bm{b}$ as
$\bm{r}_{k+1} \leftarrow \bm{r}_{k+1} - (\bm{u}_{i}^{T} \bm{r}_{k+1}) \bm{u}_{i}$
for $i=1, \dots, k$ and where $\bm{u}_{i}$ are the previous iteration's orthonormal vectors created from the residual $\bm{r}_{i}$. 
Re-orthogonalization comes at an increased memory cost, as we must store all previous residuals, adding $\mathcal{O}(JN)$ memory costs for $J$ steps of CG.
However, we find that with re-orthogonalization (and high tolerances) we are able to converge in very few steps, often $J < 50.$

\noindent \textbf{Preconditioning} \quad 
With preconditioning, we hope to find an operator such that $\bm P^{-1} \bm K \approx \bm I$ with $\bm P^{-1}(\bm v)$ is inexpensive to compute and solve the system $\bm P^{-1} \bm K \bm v = \bm P^{-1} \bm w$ instead of $\bm K \bm v = \bm w.$
Following \citet{gardner2018gpytorch}, we use the pivoted Cholesky decomposition, which requires solely access to the rows of the matrix (e.g. $\bm K_{i.} = \bm K \bm e_i$ where $e_i$ is a zero vector with $i$th entry equal to one) and an approximate diagonal (constant in our case).
Running $k$ steps of the pivoted Cholesky decomposition on $\bm K$ produces an approximation $\bm{\tilde K} \approx \bm L_k \bm L_k^\top + \sigma^2 I$ and we use the Woodbury matrix identity to construct $\bm P^{-1}$. 
\begin{align}
    \bm P^{-1} \bm w &= \sigma^{-2} \bm w - \nonumber \\
    &\sigma^{-4}\bm{L}_{k} \left(\bm{I} + \sigma^{-2} \bm{L}_{k}^{T} \bm{L}_{k}\right)^{-1} \bm{L}_{k}^{T} 
      \bm{w}.
     \label{eq:woodbury}
\end{align}
The inner system is of size $k << N$ and so can be solved using direct methods (e.g. Cholesky or QR) in single precision (due to a lack of LAPACK support for these in half).

To summarize, we show our revised CG procedure in algorithm \ref{alg:rCG},
with the differences from the BBMM CG algorithm of \citet{gardner2018gpytorch} in bolded blue font.

\begin{figure*}[!t]
\centering
\begin{subfigure}{0.19\textwidth}
\centering
\includegraphics[width=\linewidth]{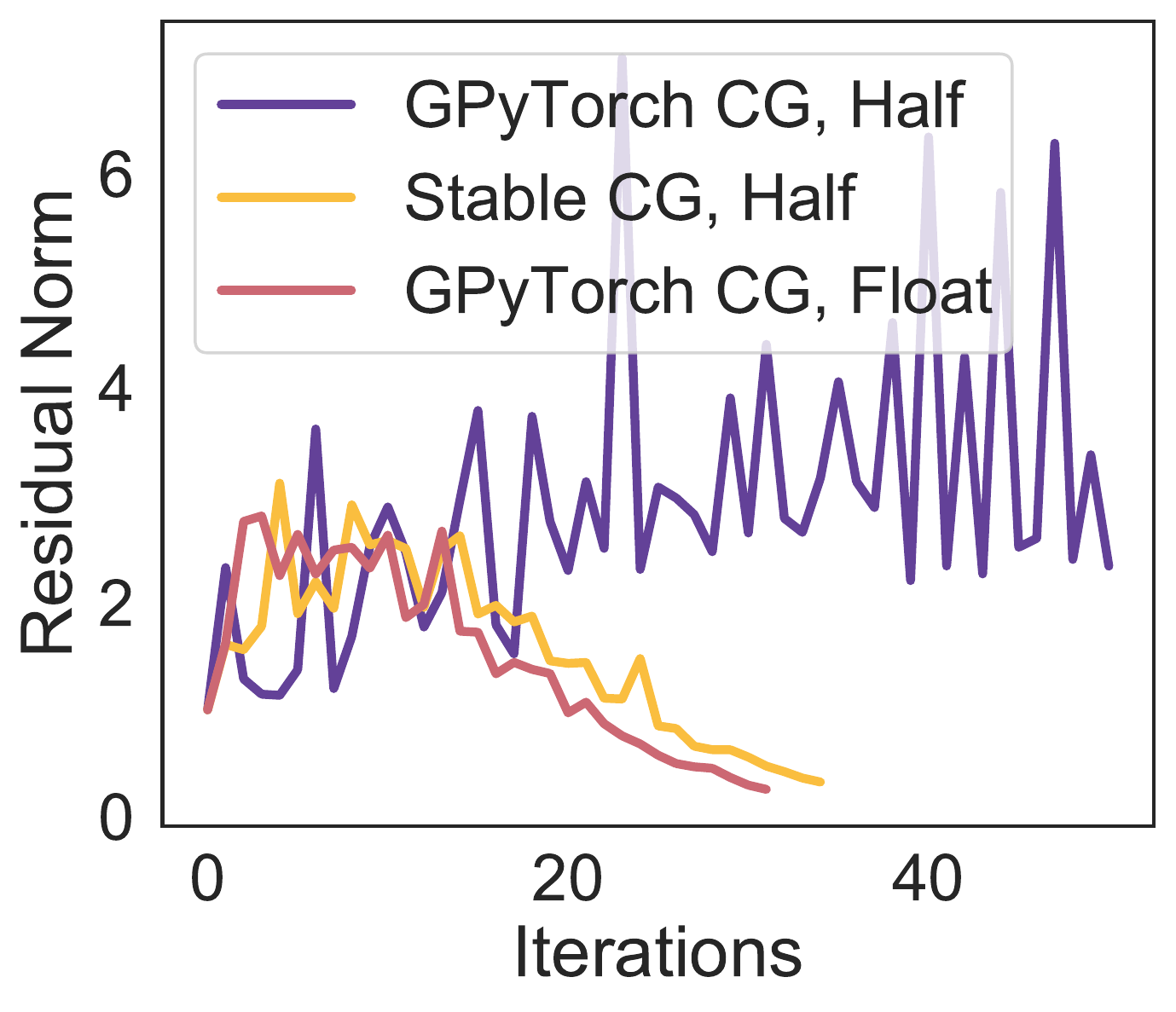}
  \caption{Elevators}
\label{fig:elevators}
\end{subfigure}
\begin{subfigure}{0.19\textwidth}
\includegraphics[width=\linewidth]{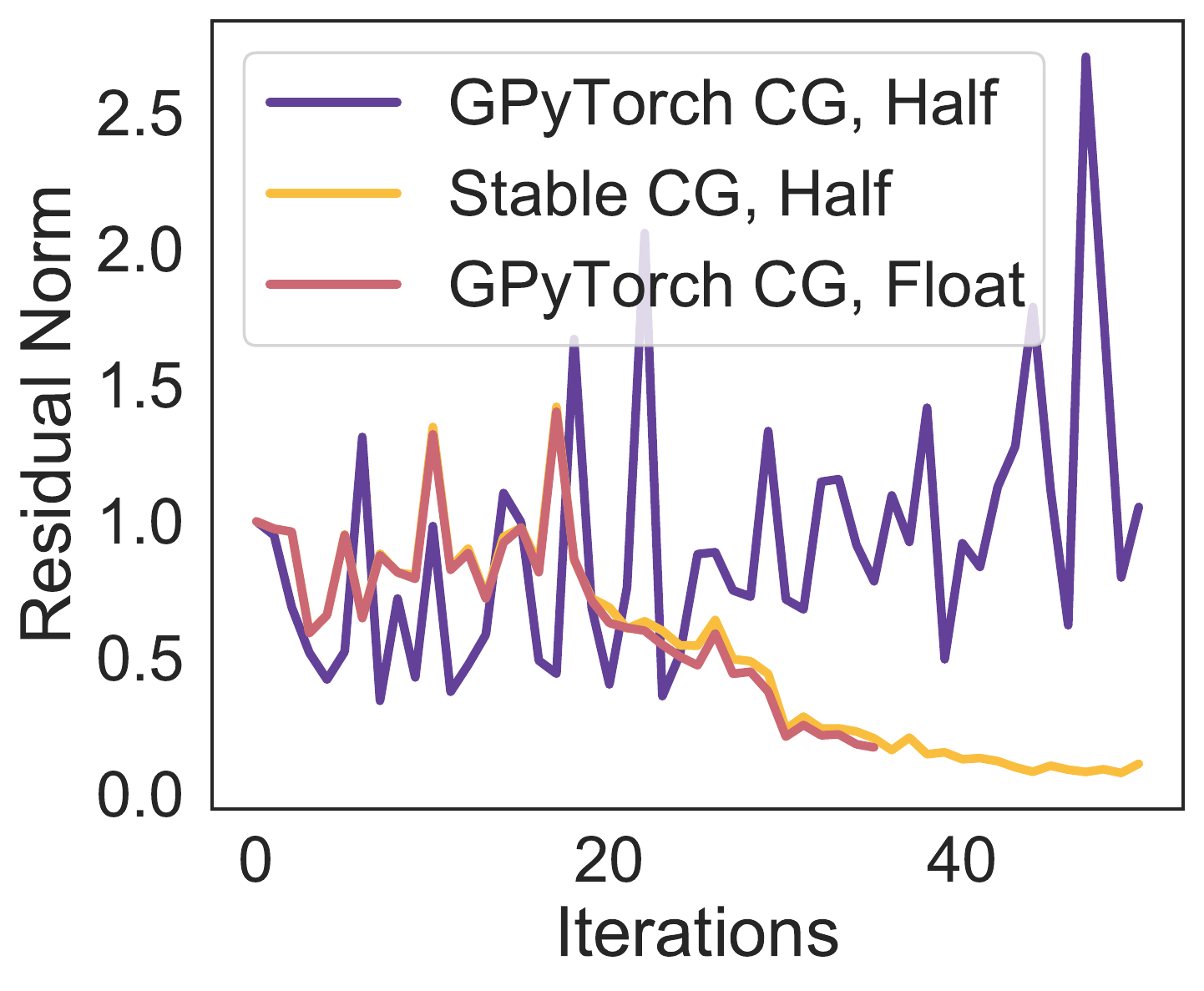}
  \caption{\emph{KeggD}}
\label{fig:keggd}
\end{subfigure}
\begin{subfigure}{0.19\textwidth}
\includegraphics[width=\linewidth]{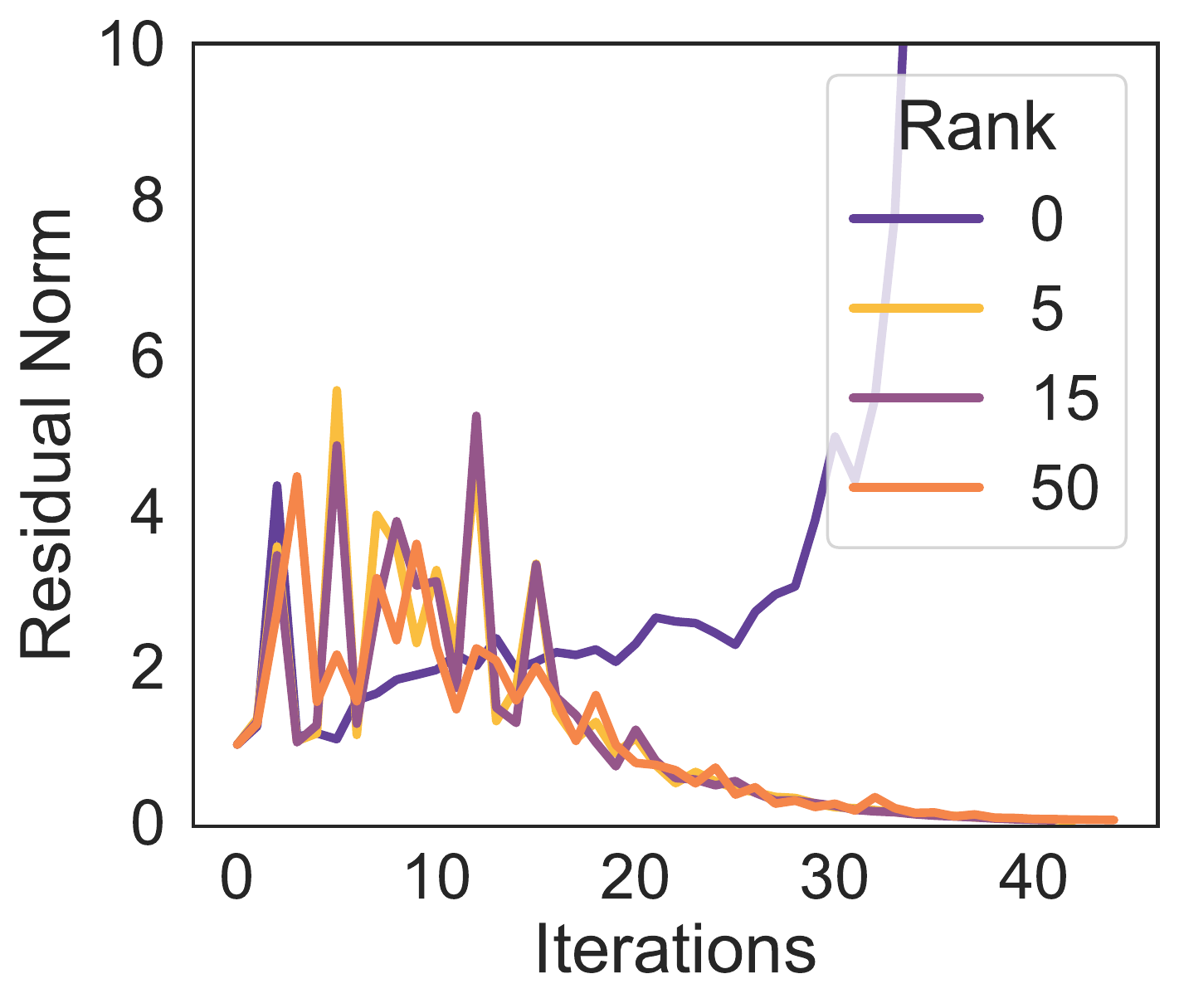}
  \caption{\emph{Buzz}}
\label{fig:buzz}
\end{subfigure}
\begin{subfigure}{0.19\textwidth}
\centering
\includegraphics[width=\linewidth]{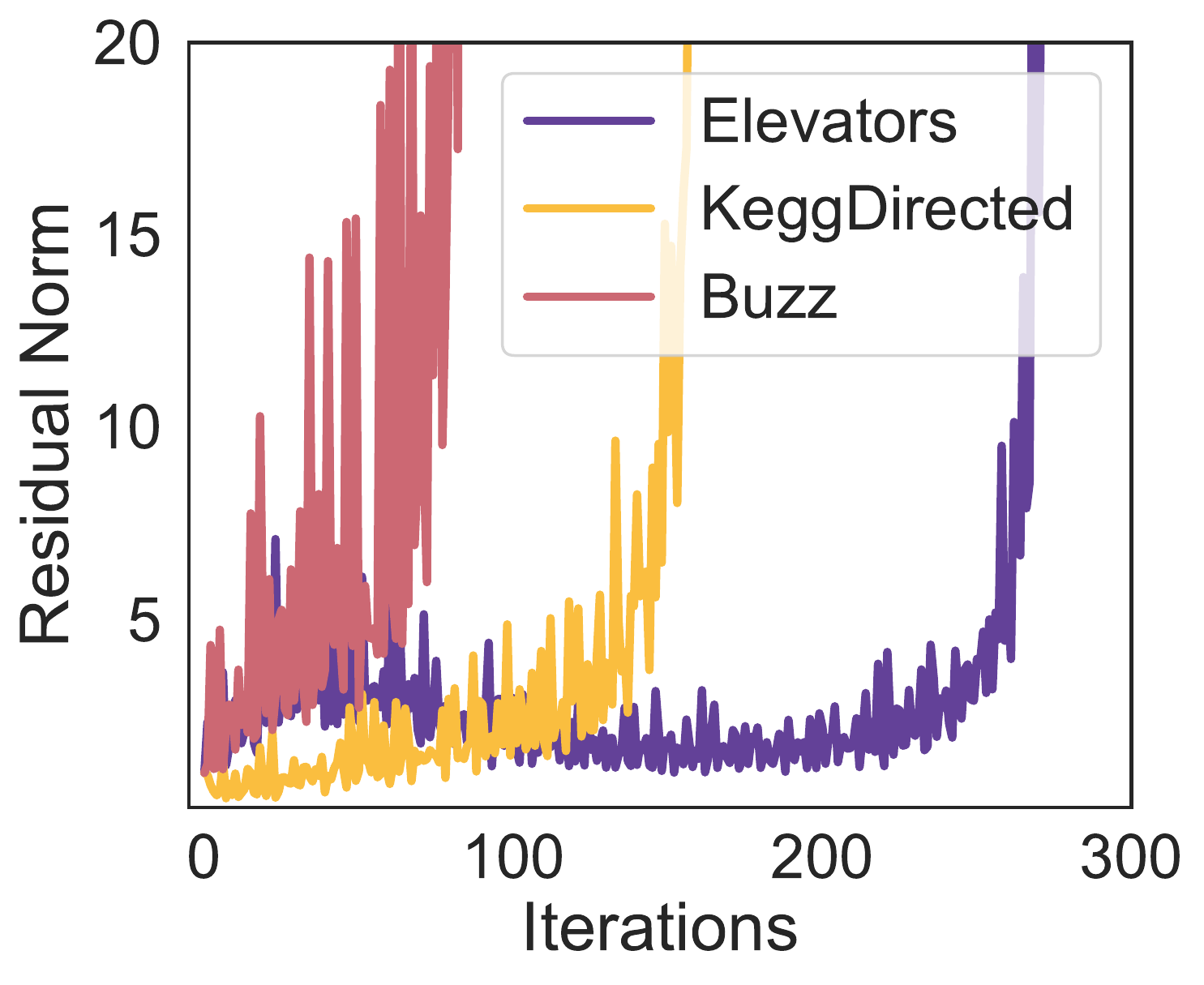}
\caption{Tolerances}
\label{fig:tol_div}
\end{subfigure}
\begin{subfigure}{0.19\textwidth}
\includegraphics[width=\linewidth]{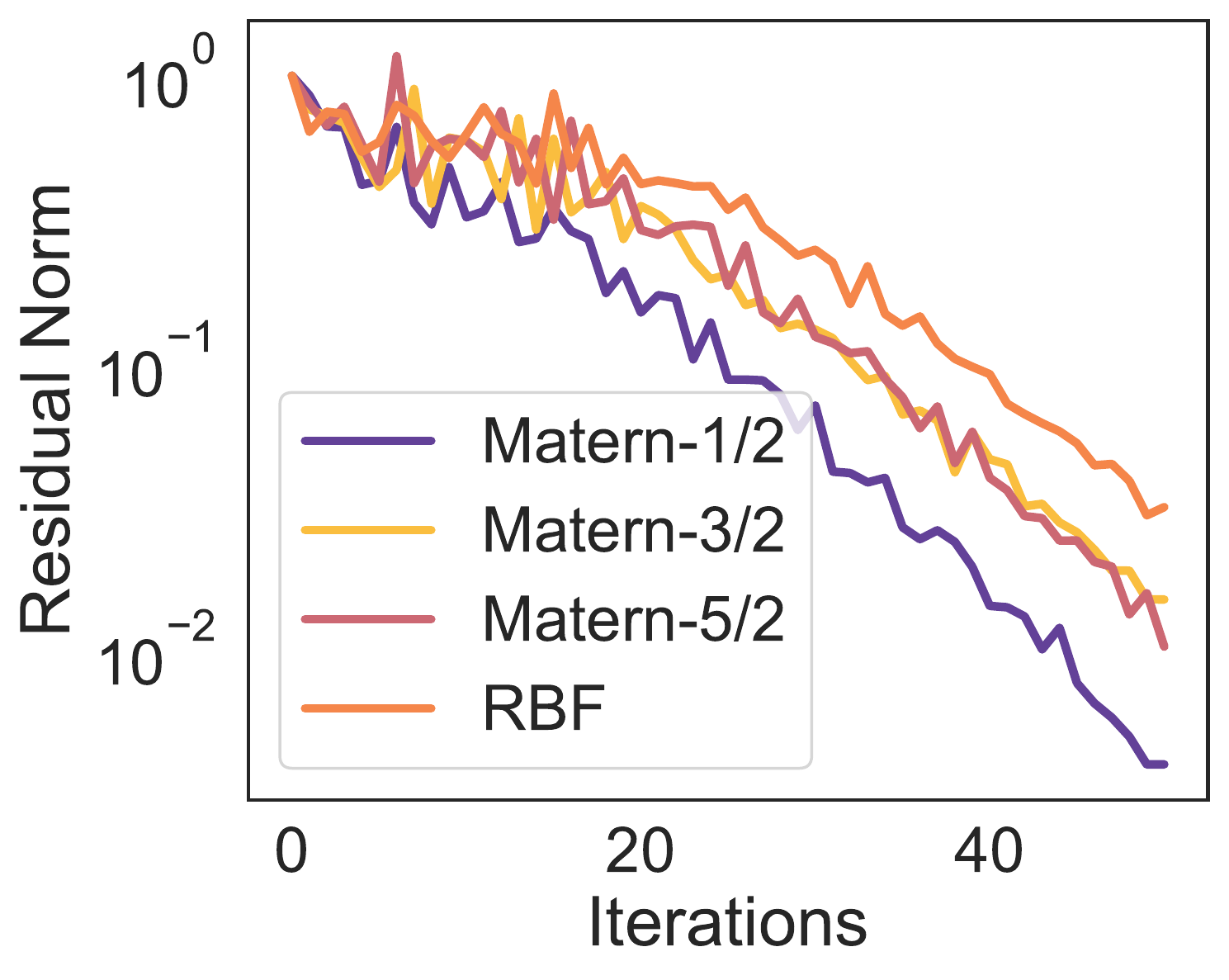}
  \caption{\emph{PoleTele}}
\label{fig:pol}
\end{subfigure}
\caption{
\textbf{(a,b)} Residual norms on \emph{Elevators} and \emph{KeggD}; standard half precision CG fails while our approach converges in a similar number of iterations to single precision CG. \textbf{(c)} Residual norms using our CG solver on \emph{Buzz}; using preconditioning prevents divergences due to round-off errors. \textbf{(d)} Running CG for too many iterations produces round-off errors that build up over time producing divergences. \textbf{(e)} Mat\'ern-$1/2,3/2$ kernels tend to converge quicker than RBF kernels on \emph{PoleTele}.
}
\label{fig:residuals}
\end{figure*}

\subsection{Loss Function}

Computation of Eq. \eqref{eq:loss} requires either an expensive eigendecomposition or Lanczos iteration in the forwards pass to compute the log determinant term, which can be quite unstable \citep{gardner2018gpytorch}.
Instead, we use a pseudo-loss that has the same gradient as Eq. \eqref{eq:gradient} via solving $\tilde K$ against both the response $\bm y$ and the probe vectors $\bm z_i$ simultaneously. 
That is, we find solutions to the linear system using Alg. \ref{alg:rCG}:
\begin{equation} \label{eq:cg}
    \begin{split}
      \bm{K} \left[\bm{u}_{0}, \bm{u}_{1}, \cdots, \bm{u}_{M}\right]
      =
      \left[\bm{y}, \bm{z}_{1}, \cdots, \bm{z}_{M}\right].
    \end{split}
\end{equation}
for $\bm{u}_i.$
We then detach these solutions and compute simply:
\begin{equation}\label{eq:approx_loss}
    \tilde{\mathcal{L}}(\bm{\theta})
      = \frac{1}{2 M} \sum_{j=1}^{M}  \bm{u}_{j}^{T} \left(\bm{K}_\bm{\theta} \bm{z}_{j}\right) - \frac{1}{2} \bm{u}_{0}^{T} \left(\bm{K}_\bm{\theta} \bm{u}_{0}\right),
\end{equation}
which has gradient
\begin{equation} \label{eq:approx}
    \begin{split}
      \nabla_{\bm{\theta}} \tilde{\mathcal{L}}
      = \frac{1}{2 M} \sum_{j=1}^{M}  \bm{u}_{j}^{T} \left(\nabla_{\bm{\theta}} \bm{K} \bm{z}_{j}\right) - \frac{1}{2} \bm{u}_{0}^{T} \left(\nabla_{\bm{\theta}} \bm{K} \bm{u}_{0}\right),
    \end{split}
\end{equation}
which is the same as Eq. \ref{eq:gradient}.
Eq. \ref{eq:approx} then only requires gradients with respect to matrix vector products, which are computed natively in KeOps.
Eq. \eqref{eq:approx} does not use variance reduction strategies such as \citet{wenger2021reducing} but they could also be incorporated.

The computational complexity of these operations is the same as that of computing the standard log marginal likelihood \citep{gardner2018gpytorch}.
Computing Eq. \ref{eq:approx} requires $\mathcal{O}((M+1)N^2)$ time where $M$ is the number of trace estimates and $N$ is the total number of data points.
Using half precision does not alter the
order of the cost but only reduces the constant 
of the MVMs by half.

\section{Benchmarking Experiments}\label{sec:hp_bencharmsk}

\begin{table*}[!ht]
\caption{RMSEs and training time with $\pm$ the standard deviation of the results over $5$ different seeds on a suite of UCI tasks for half and single precision GPs and SVGPs. Here, we use RBF ARD kernels with $50$ CG iterations and $50$ optimization steps. We included KeOps compilation times and observe that the largest improvements come on \emph{Song} and \emph{HouseElectric}.}
\label{tab:rbf_ard}
\centering
\scriptsize{
\begin{tabular}{ccccccccc}
  \midrule
  \multicolumn{2}{c}{\textbf{}} & 
  \multicolumn{3}{c}{\textbf{RMSE}} & 
  \multicolumn{2}{c}{\textbf{Time}} & 
  \\
  \multicolumn{1}{c}{\textbf{Dataset}} & 
  \multicolumn{1}{c}{$(N , d)$} & 
  \multicolumn{1}{c}{\textbf{Single}} & 
  \multicolumn{1}{c}{\textbf{Half}} &
  \multicolumn{1}{c}{\textbf{SVGP}} &
  \multicolumn{1}{c}{\textbf{Single}} & 
  \multicolumn{1}{c}{\textbf{Half}} &
  \multicolumn{1}{c}{\textbf{SVGP}} &
  \\ \hline
  PoleTele & (13.5K, 26) 
  &  $0.11 7 \pm 0.004$ & $0.121 \pm 0.003$ & $0.14 \pm 0.004$ & $57.6 \pm 0.6$ & 88.3 $\pm$ 3.2 & $15. \pm 0.5$ \\
  Elevators & (14.9K, 18) 
  & $0.364 \pm 0.004$ & 0.382 $\pm$ 0.001 & $0.374 \pm 0.005$ & $58.4 \pm 3.1$ & 90.1 $\pm$ 2.9 & $16 \pm 0.5$ \\
  Bike & (15.6K, 17) 
  & $0.074 \pm 0.003$ & 0.083 $\pm$ 0.009 & $0.077 \pm 0.006$ & $62.3 \pm 0.7$ & 85.6 $\pm$ 0.41 & $17 \pm 0.6$ \\
  Kin40K &(36K, 8) 
  & $0.099 \pm 0.001$ & 0.100 $\pm$ 0.003 & $0.165 \pm 0.003$ & $65.5 \pm 5.3$ & 92.9 $\pm$ 0.28 & $39 \pm 0.5$\\
  Protein & (41.1K, 9) 
  & $0.055 \pm 0.006$ & 0.635 $\pm$ 0.002 & $0.632 \pm 0.01$ & $70.9 \pm 5.7$ & 115.7 $\pm$ 0.10 & $46 \pm 3.2$ \\
  3droad & (391.4K, 3) 
  & $0.194 \pm 0.01$ & 0.215 $\pm$ 0.004 & $0.412 \pm 0.011$ & $1,260 \pm 35$ & 1,003 $\pm$ 0.97 & $412 \pm 101$ \\
  Song & (463.8K, 90) 
  & $0.761 \pm 0.004$ & 0.779 $\pm$ 0.022 & $0.999 \pm 0.002$ & 24,357 $\pm$ 2,613 & 9,930 $\pm$ 130 & $284 \pm 2.0$ \\
  Buzz & (524.9K, 77) 
  & $0.300 \pm 0.01$ & 0.448 $\pm$ 0.025 & $0.268 \pm 0.004$ & 25,436 $\pm$ 1,200 & 23,127 $\pm$ 1,819 & $617 \pm 114$\\
  HouseElectric & (1844.3K, 9) 
  & $0.052 \pm 0.002$ & 0.051 $\pm$ 0.004 & - & 42,751 $\pm$ 2,180 & 36,025 $\pm$ 1,178 & -  \\\midrule
\end{tabular}
}
\end{table*}

We compare our method directly to single precision GP training using KeOps and GPyTorch on benchmarks of up to $1.8$ million data points from the UCI repository \citep{Dua:2019} using RBF ARD kernels and a single NVIDIA V100 GPU. We show results for other kernels in Appendix \ref{sec:exp_details}.
A mixed precision strategy of using solely half precision matrix vector multiplications produces speedups of at least two times over an equivalent single precision model.

\paragraph{Residual Norms}
First, we consider the relative solve error on three different datasets of varying size: \emph{Elevators}, \emph{KeggDirected}, and \emph{Buzz} for hyperparameters collected at the end of the optimization runs. While standard CG in half precision diverges very quickly, we find that our stable CG implementation in half precision closely matches the convergence behaviour of single precision CG, converging to a residual tolerance less than $0.5$ in less than $50$ optimization steps, as shown in Figures \ref{fig:elevators} for \emph{elevators} and \ref{fig:keggd} for \emph{KeggDirected}.

Furthermore, we explore the effects of using no preconditioner (rank $0$) and also of using
preconditioners of rank $5$, $15$ and $50$ on \emph{Buzz}.
As seen in Figure \ref{fig:buzz}, we find that CG diverges without preconditioning and that
a rank 5 preconditioner is sufficient for convergence and behaves similarly to preconditioners with rank 15 and 50. These findings are consistent with \citet{maddox2021iterative}.

Despite preconditioning, the effect of low-precision does produce round-off errors that grow with the number of iterations, 
preventing us from running our method for a large number iterations or with low tolerances.
We show this effect across three datasets in Figure \ref{fig:tol_div}.
Finally, we show the results across Mat\'ern kernels with varying $\nu$ and RBF kernels without ARD on \emph{PoleTele} in Figure \ref{fig:pol}, finding that Mat\'ern-$1/2$ kernels tend to converge fastest. This is consistent with our discussion in Section \ref{sec:eigenspectrum}, 
since the Mat\'ern-$1/2$ kernels have a wider range of numerically representable distances, reducing round-off error and they also have better conditioning.

\paragraph{Optimization Trajectories}

\begin{figure}[t]
    \centering
        \includegraphics[width=0.9\linewidth]{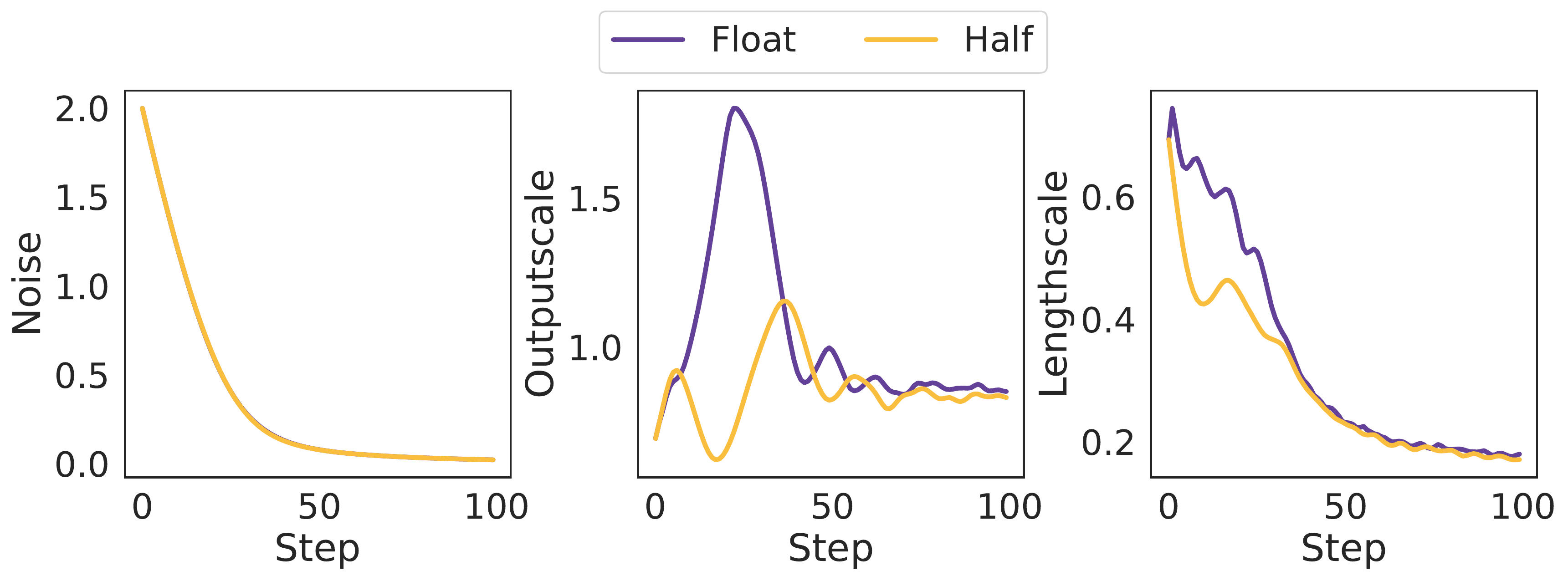}
    \caption{Optimization and parameter trajectories on \emph{Protein} for both a standard single precision model and our half precision implementation. The noise term (left) and lengthscale (right) are very close for both precisions, while the outputscale (center) shows some minor differences, but convergence to a similar value.}
    \label{fig:opt_trajectories}    
\end{figure}
Next, in Figure \ref{fig:opt_trajectories}, we display the evolution of the three hyperparameters (noise, outputscale, and lengthscale) on \emph{Protein} with the same plot shown for \emph{3droad} in Figure \ref{fig:3droad_comp}. 
We see that the solver produces accurate enough solves so that gradients are unaffected and the hyperparameters train similarly to the GPyTorch solves in single precision. 

\paragraph{Benchmark Results}

\begin{figure}[!ht]
      \centering
      \includegraphics[width=7cm,height=5cm]{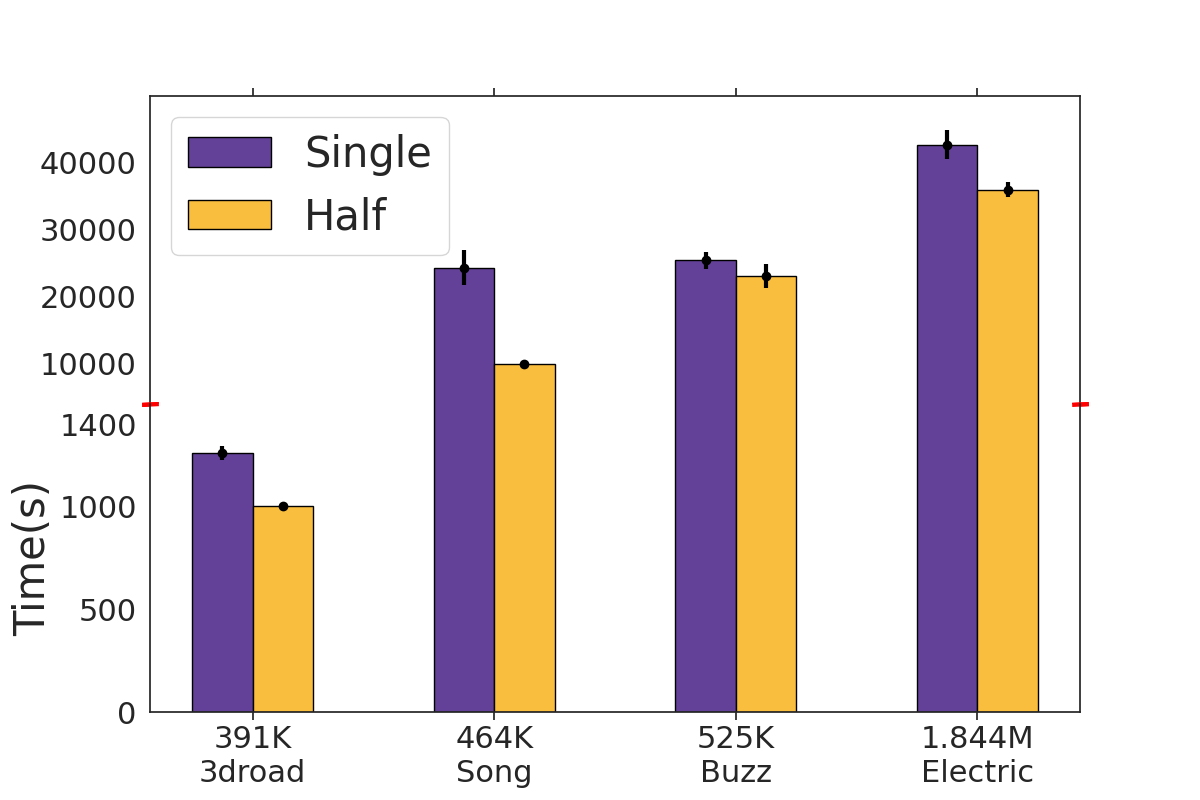}
      \caption{Half precision GP inference is particularly faster than full precision on large datasets.
      Running times ($\pm$ one standard deviation) for the experiments of Table~\ref{tab:rbf_ard} with the largest datasets.
      The scale of the y-axis is broken into [0-1,500] and [5,000-50,000].}
      \label{fig:small_without_comp}
    \end{figure}

Finally, in Table \ref{tab:rbf_ard} we show the root mean square errors (RMSEs) and fitting times for both single and half precision as well as variational GPs (SVGPs) \citep{hensman2013gaussian} over $5$ seeds, displaying the mean and standard deviation. 
At a first glance, it appears that half precision only runs faster for datasets with size larger than $100$K datapoints.
However, this is simply because the fixed compilation times in KeOps for half are longer, and the runtime is already very fast for smaller datasizes.
As seen in Figure \ref{fig:without_comp}, when taking out the compilation times, half precision always runs faster.
Moreover, for \emph{Song} and \emph{HouseElectric}, we see significant speedups (of up to $3$x) for the same number of optimization steps
as seen in Figure \ref{fig:small_without_comp}.
Finally, SVGPs tend to be faster but often perform significantly worse than half or single precision GPs.
We were unable to run SVGPs on \emph{HouseElectric} due to out of memory errors at test time on our servers. Combining half-precision
inference with SVGP is an interesting direction for future work.

\section{Conclusion}

Deep learning has benefited enormously from advances in hardware design and systems. While GPs have started to exploit GPU parallelization, with highly impactful systems such as GPyTorch \citep{gardner2018gpytorch}, there is essentially no work on low-precision inference with GPs, despite the widespread impact of low-precision optimization in deep learning. Indeed, low-precision GPs are nontrivial because the operations necessary for inference are relatively sensitive to the precision of the computations. In this respect, our work provides several new insights:
\begin{itemize}
    \item \emph{Training GPs in low-precision requires algorithmic modifications to ensure numerical stability}. To this end, we propose a stable version of conjugate gradients that uses scale translation, re-orthogonalization, preconditioning and mixed precision.
    \item \emph{Mixing the precision of computations guarantees speedups while minimizing issues related to round-off errors}. To minimize catastrophic round-off error it is important to cast inexpensive but critical computations, such as the step sizes in CG, into higher precisions, while performing noncritical and large computations, such as MVMs, on lower precisions to gain runtime speedups.
    \item \emph{The kernel choice affects the convergence behavior of GPs in lower precisions}. M\'{a}tern kernels converge faster than RBF kernels,
    as M\'{a}tern kernels have a wider range of numerically representable distances, reducing round-off error and poor conditioning.
    \item \emph{Lower precisions slow the rate of decay of the eigenspectrum of the kernel matrices}. 
    The change in the rate of decay of the eigenvalues introduced by using lower precisions has a detrimental effect both on the training convergence, by slowing the rate of progress of the CG iterations,
    and on the provable generalization guarantees, by increasing the effective dimensionality.
    \item \emph{How the bits are split between matissa and exponent affect the training results}. We find float16 to be the easiest half precision standard to work with since 
    it has a wider range of values (when compared to bfloat16) which avoids the CG steps from being clipped.
    \item \emph{Our algorithm for training GPs in lower precisions can be used to broadly accelerate standard CG, beyond Gaussian process inference}. In many cases, there is no reason not to use the proposed procedures as a drop-in replacement for standard CG. Apart from re-orthogonalization, the rest of the modifications that we introduced into CG have no effect on single or double precision, and apply broadly.
\end{itemize}

These contributions will continue to appreciate with time, as hardware advances continue to further accelerate low-precision computations. 

There are also many promising directions for future research. The use of lower precisions
and our mixed precision approach can be used in other iterative methods such as  Lanczos \citep{pleiss2018constant} and MINRES \citep{pleiss2020fast}. Our methods could also be adapted to approximate GP inference, e.g., for classification, and combined with SVGP and scalable low rank approximations of the kernel matrix.

\paragraph{Acknowledgements} \quad
We would like to thank Chris De Sa and Ke Alexander Wang for helpful discussions. This research is supported by an Amazon Research Award, Facebook
Research, Google Research, Capital One,  NSF CAREER IIS-2145492, NSF I-DISRE 193471, NIH R01DA048764-01A1, NSF IIS-1910266, and NSF 1922658 NRT-HDR.

\bibliography{ref.bib}

	\onecolumn
	\appendix
	\renewcommand\thefigure{A.\arabic{figure}}
	\renewcommand\thetable{A.\arabic{table}}
	\setcounter{figure}{0}

	Our Appendix is structured as follows: 
	\begin{itemize}
		\item In Appendix \ref{app:relwork}, we further describe related work, including on conjugate gradients.
		\item In Appendix \ref{app:experiments}, we show several other experiments on both the properties of half precision kernel matrices and half precision conjugate gradients.
		\item In Appendix \ref{sec:exp_details}, we outline experimental details for all of our experiments.
		\item In Appendix \ref{app:theory}, we give some detailed theoretical analysis of half precision kernel matrices, focusing on the quantized effective dimension and the effect of finite precision on the support of the kernel.
	\end{itemize}
	
	\section{Extended Related Work}\label{app:relwork}
	
	\paragraph{Conjugate Gradients: } 
	A description of the conjugate gradients algorithm is given in Alg. \ref{alg:CG_basic} while using preconditioning \citep{nocedal2006,golub2018matrix}.
	\citet{gardner2018gpytorch} propose a variant of conjugate gradients that they call modified batched CG (mBCG) which we use in our work. 
	The primary difference between mBCG and CG is that mBCG enables solving several linear systems at once by performing all computations in batch mode so that linear operators such as $\bm K (\bm v)$ are actually matrix matrix multiplications rather than matrix vector products.
	Then, an individual set of learning rates $\alpha_k$ and $\beta_k$ is used for each system.
	Our stable CG implementation (Alg. \ref{alg:rCG}) is actually based off of mBCG, but for didactic purposes we display only the standard CG version.
	
	\begin{algorithm}[!ht]
		\caption{CG }
		\label{alg:CG_basic}
		\begin{algorithmic}[1]
			\STATE \textbf{Input:} MVM function $\bm{K}\left(\cdot\right)$, initial solution guess $\bm{x}_{0}$, linear system right hand side $\bm{b}$, 
			tolerance $\epsilon$, preconditioner function $\bm{P}\left(\cdot\right)$
			\STATE \textbf{Initialize:} $k \leftarrow 0$, $\bm{r}_{0} \leftarrow \bm{K}\left(x_{0}\right) - \bm{b}$, $\bm{d}_{0} \leftarrow -\bm{r}_{0}$,
			$\bm{z}_{0} = \bm{P}\left(\bm{r}_{0}\right)$ and
			$\gamma_{0} = \bm{r}_{0}^{T} \bm{z}_{0}$.
			\vspace{0.5em}
			\WHILE{$\norm{\bm{r}_{k}}_{2} < \epsilon$}
			\STATE $\alpha_{k} = \frac{\gamma_{k}}{\bm{d}_{k}^{T} \bm{K}\left(\bm{d}_{k}\right)}$
			\STATE $\bm{x}_{k+1} = \bm{x}_{k} + \alpha_{k} \bm{d}_{k}$
			\STATE $\bm{r}_{k+1} = \bm{r}_{k} + \alpha_{k} \bm{K}\left(\bm{d}_{k}\right)$
			\STATE $\bm{z}_{k+1} = \bm{P}\left(\bm{r}_{k+1}\right)$
			\STATE $\gamma_{k+1} = \bm{r}_{k+1}^{T} \bm{z}_{k+1}$
			\STATE $\beta_{k+1} = \frac{\gamma_{k+1}}{\gamma_{k}}$
			\STATE $\bm{d}_{k+1} = -\bm{r}_{k+1} + \beta_{k+1} \bm{d}_{k}$
			\ENDWHILE
		\end{algorithmic}
	\end{algorithm}
	
	\paragraph{Other Scalable Gaussian Processes}
	
	We note that our matrix-free schemes can be used to scale up approximate kernel methods such as Nystrom style approximations \citep{smola2000sparse,williams2000using}; 
	indeed, \citet{meanti2020kernel} use Nystrom approximations and KeOps for their kernel ridge regression approach.
	However, \citet{pmlr-v89-zhang19f} found limited speedups when quantizing (which is slightly distinct from half precision) Nystrom approximations.
	Similarly, \citet{pleiss2020fast} used iterative methods (in their cast MINRES) to speed up variational Gaussian processes \citep{titsias2009variational,hensman2013gaussian}, and we hope to speed up their approach as well.
	\citet{chen2013parallel,nguyen2019exact} proposed parallel direct Cholesky based GP schemes for more scalable GP regression; however, their approaches will probably perform poorly in lower precision, as we demonstrate is the case for pivoted Cholesky based solves in Section \ref{sec:pc_exp}.
	
	Finally, kernels with compact support have been previously studied from a kernel approximation point of view \citep{genton2001classes,gneiting2002compactly}.
	However, these works focused on developing new techniques to approximate an infinitely supported kernel with a kernel that has demonstrated compact support, rather than using floating point precision to develop an approximate kernel with compact support.
	
	\section{Extended Experiments}\label{app:experiments}
	
	\paragraph{Summation Approaches}
	
\begin{figure}[!ht]
		\centering
		\begin{subfigure}{0.32\textwidth}
			\centering
			\includegraphics[height=3.5cm]{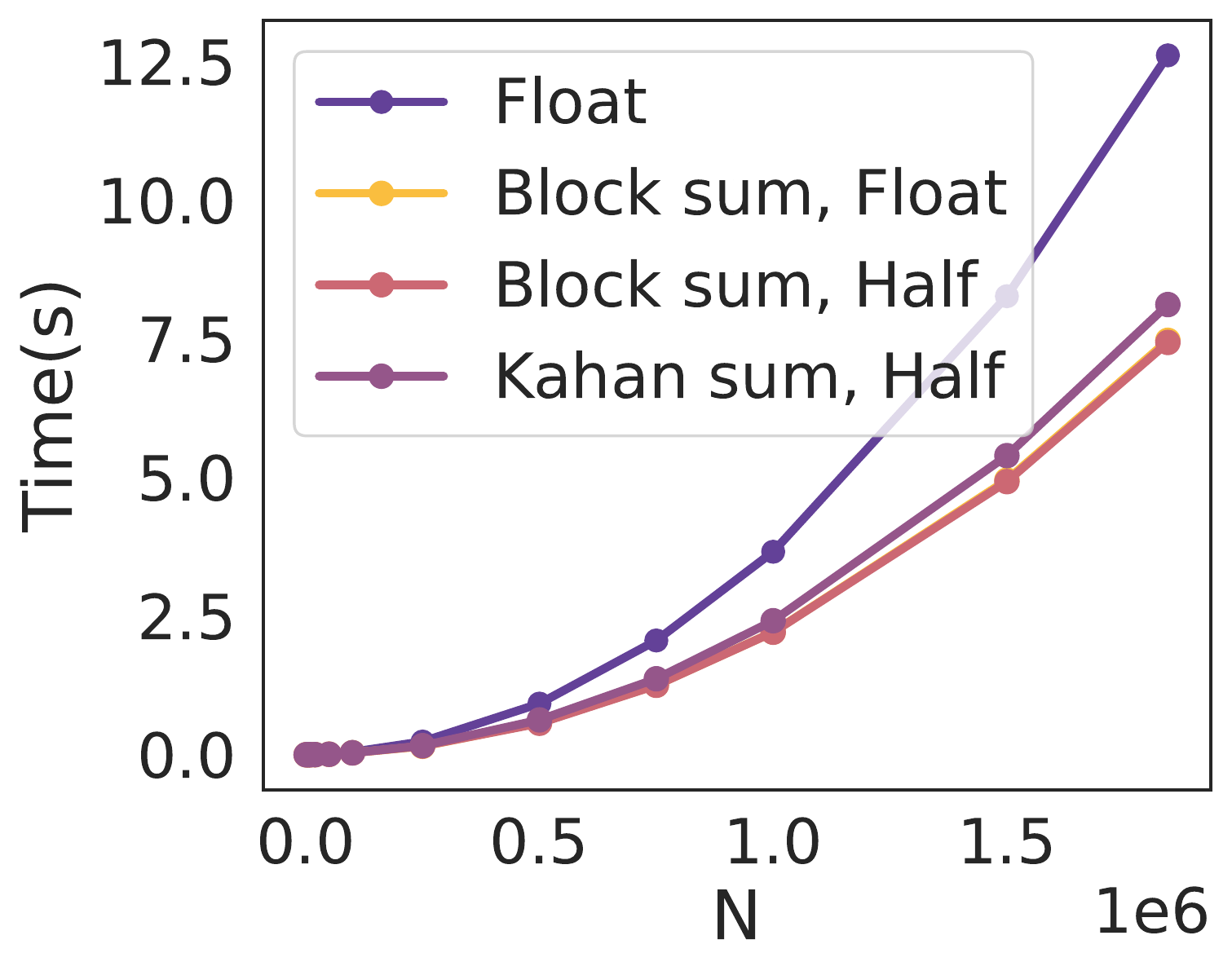}
			\caption{Summations.}
			\label{fig:mm_full_times}
		\end{subfigure}
		\begin{subfigure}{0.32\textwidth}
			\centering
			\includegraphics[height=3.5cm]{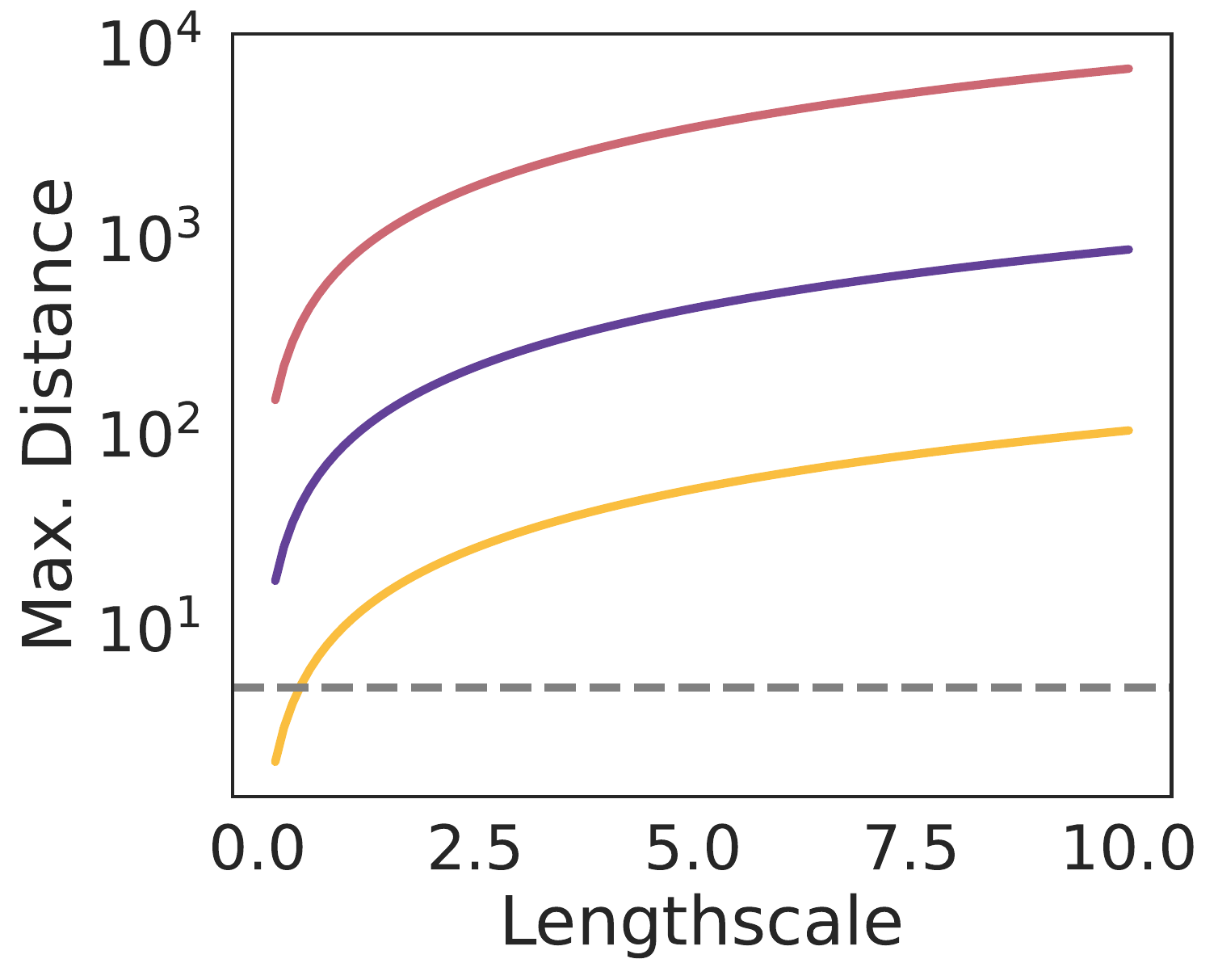}
			\caption{Mat\'ern-$1/2.$}
			\label{fig:max_distances_matern}
		\end{subfigure}
		\begin{subfigure}{0.32\textwidth}
			\centering
			\includegraphics[height=3.5cm]{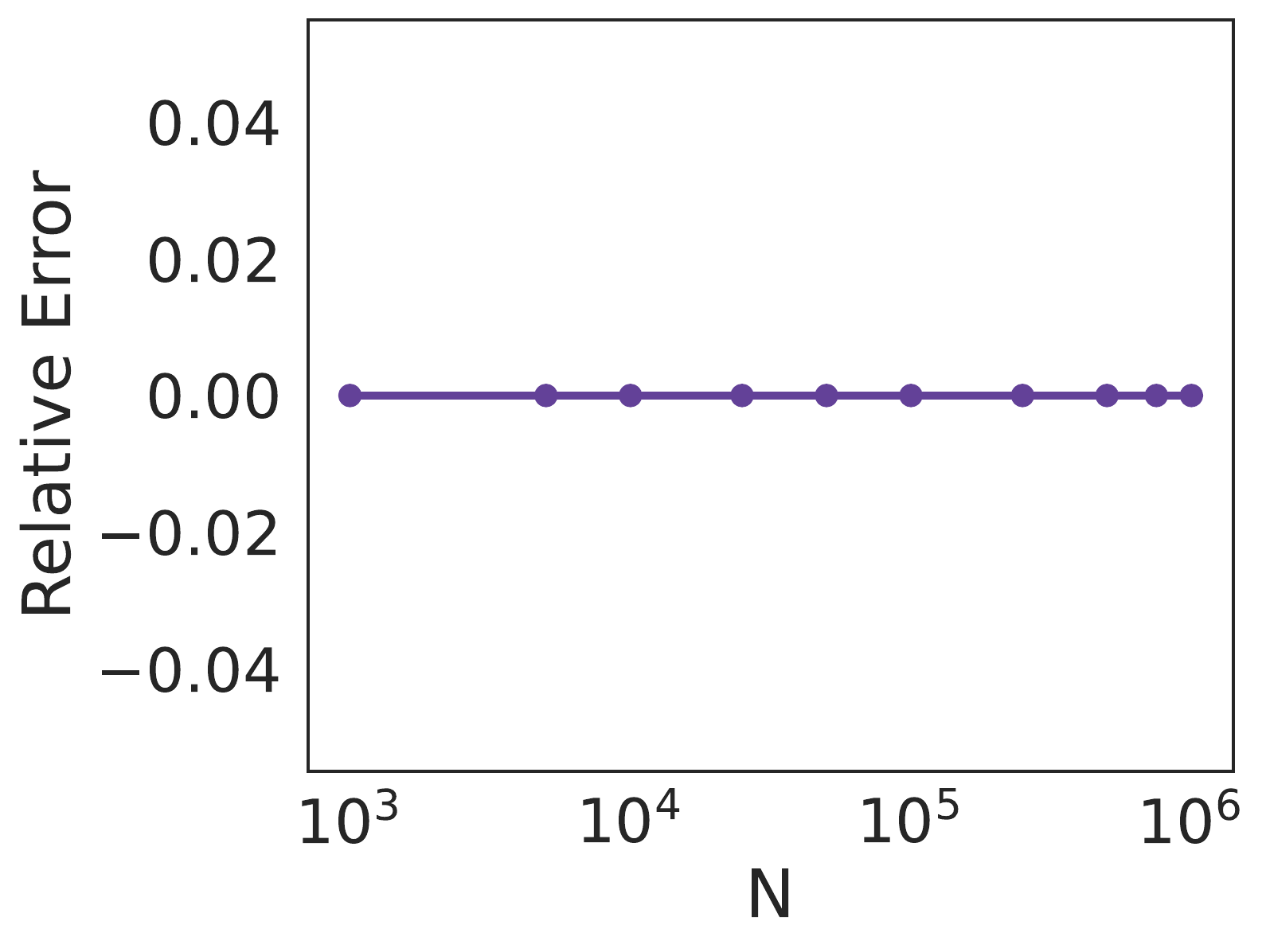}
			\caption{Error of sparsified MVMs.}
			\label{fig:sparsity_error}
		\end{subfigure}
		\caption{\textbf{(a)} Block summation in floating point adds negligible overhead compared with block summation in half precision, while being as accurate as Kahan summation.
			\textbf{(b)} Maximum distance representable for Mat\'ern-$1/2$ kernel; note the similar trend to Figure \ref{fig:max_distances_rbf}.
			\textbf{(c)} Error of the truncated MVM is zero as expected. To produce the sparsified MVM, we truncated any data points that had kernel entries that were un-representable in half precision.
		}
	\end{figure}

	In Figure \ref{fig:mm_full_times}, we display the different times of block summation across precisions, as well as Kahan summation and floating point accumultion of float kernel matrix MVMs, finding that all half precision accumulation mechanisms behave similarly, with Kahan summation being slightly slower than the other two.
	This plot is inspired by the study performed by the authors of KeOps \citep{charlier2021kernel}, available at \url{https://www.kernel-operations.io/keops/_auto_benchmarks/plot_accuracy.html}.
	Due to these results, we use block summation, casting each block's summation up to float before down-casting to half, as is the default in KeOps.
	
	\begin{figure}
		\centering
		\begin{subfigure}{0.32\textwidth}
			\centering
			\includegraphics[height=3.5cm]{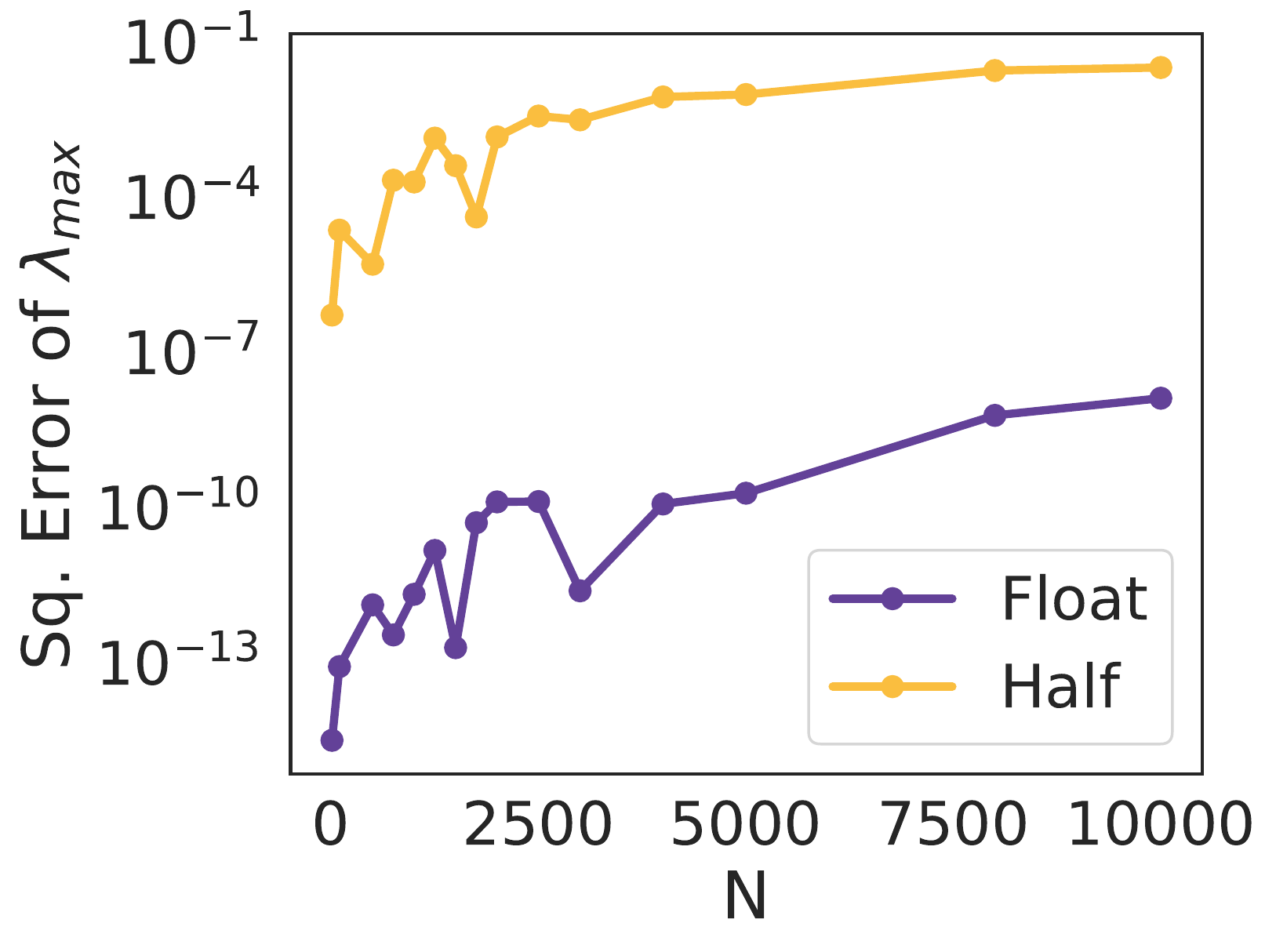}
			\caption{Difference of $\lambda_{max}.$}
			\label{fig:eval_diff}
		\end{subfigure}
		\begin{subfigure}{0.32\textwidth}
			\centering
			\includegraphics[height=3.5cm]{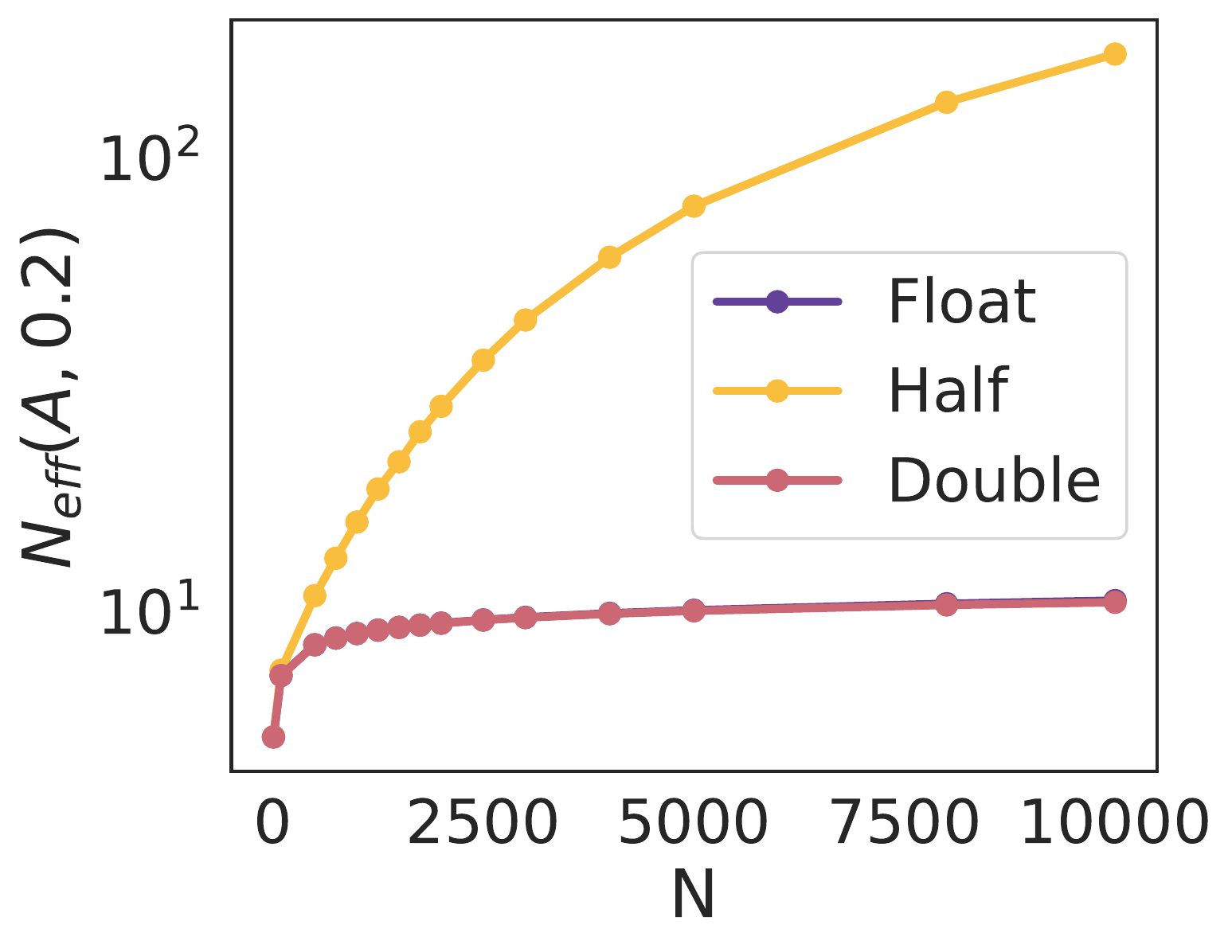}
			\caption{ED, RBF.}
		\end{subfigure}
		\begin{subfigure}{0.32\textwidth}
			\centering
			\includegraphics[height=3.5cm]{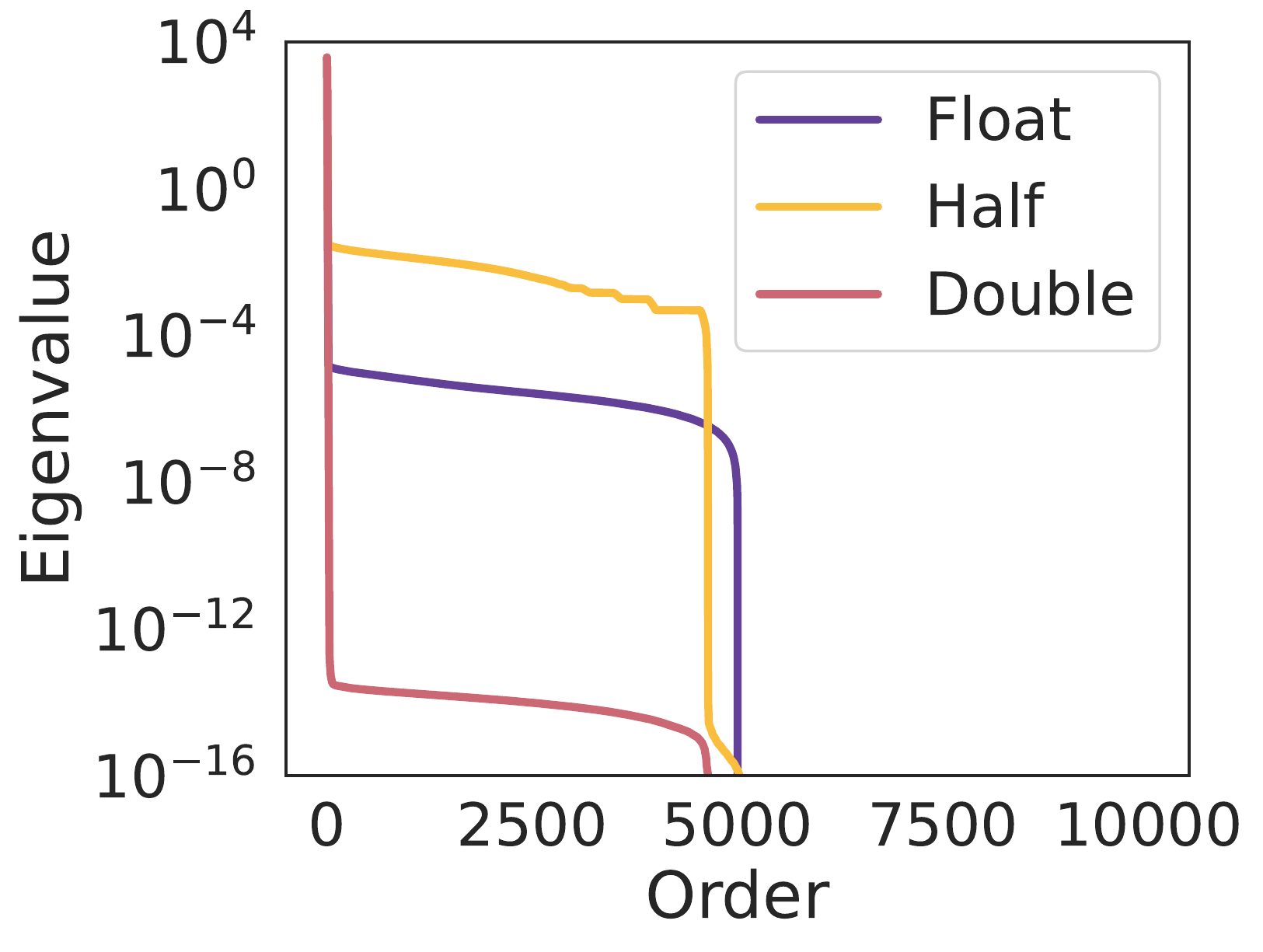}
			\caption{Spectrum, RBF.}
		\end{subfigure}
		\caption{
			\textbf{(a)} Difference of $\lambda_{\text{max}}$ across precisions for Mat\'ern-$1/2$ kernel. Other kernels have similar eigenvalue diffeences.
			\textbf{(b)} Effective dimension (ED) for RBF kernels, the trend is similar to that of the Mat\'ern-$1/2$ kernel because the eigenvalue spectrum \textbf{(c)} has a similar bunched up pattern in half.}
		\label{fig:other_spectra}
	\end{figure}
	
	\begin{figure}[!ht]
		\centering
		\includegraphics[width=0.66\linewidth]{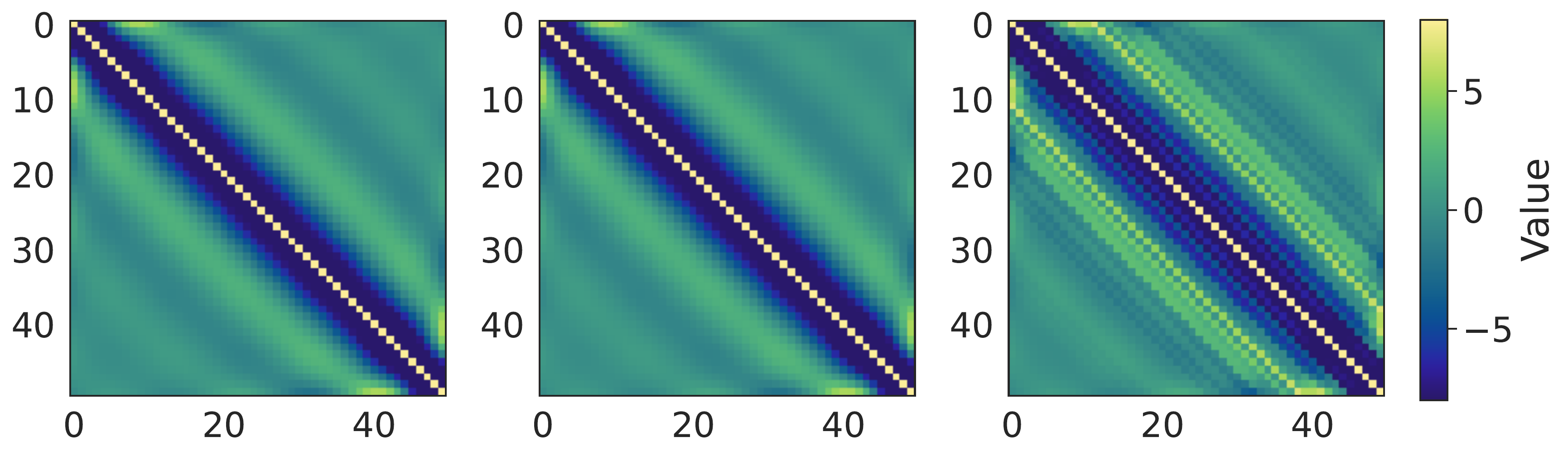}
		\caption{Matrix inverses of an RBF kernel in double (left), float (middle), and half (right) precisions. The inverse is performed in double precision, while the evaluation itself is performed in half precision. The half precision inverse is qualitatively distinct from the other two indicating a slightly distinct spectrum.}
		\label{fig:rbf_kernels_inv}
	\end{figure}
	
	\paragraph{Properties of Half Precision Kernel Matrices}
	
	In Figure \ref{fig:max_distances_matern}, we display the maximum distance representable for Mat\'ern-$1/2$ kernels across varying lengthscales, as we show for RBF and rational quadratic kernels in Figures \ref{fig:max_distances_rbf} and \ref{fig:max_distances_rq}.
	The trend for the Mat\'ern family is similar to that of the RBF kernels, except that larger distances are representable.
	
	Finally, in Figure \ref{fig:sparsity_error}, we show the error of sparsified MVMs (which is zero) across increasing dataset size for the data reduction experiment in Section \ref{sec:support}.
	
	The difference of the largest eigenvalue of a Mat\'ern-$1/2$ kernel is shown in Figure \ref{fig:eval_diff} in float and half as compared to double precision (which we use as a proxy for infinite precision).
	Note that extremely small relative differences for these largest values.
	
	In Figure \ref{fig:other_spectra}\textbf{(b)}, we show ED for RBF kernels with the associated spectrum in \textbf{(c)}.
	
	In Figure \ref{fig:rbf_kernels_inv}, we display $(\bm K + 0.01)^{-1}$ for RBF kernels with lengthscale $1$ and $50$ data points in $[-3, 3]$ across double (left), float (middle), and half (right) precisions.
	We first evaluate the kernel to a lower precision and then pass into double precision before using a Cholesky factorization to invert the kernel matrix, finding that the half precision kernel inverse has a distinct pattern (larger magnitude off-diagonal values) compared to the float and double inverse matrices.
	
	\paragraph{Benchmarking Half Precision CG}
	
	\begin{figure}[ht]
		\centering
		\begin{subfigure}{0.23\textwidth}
			\includegraphics[width=\linewidth]{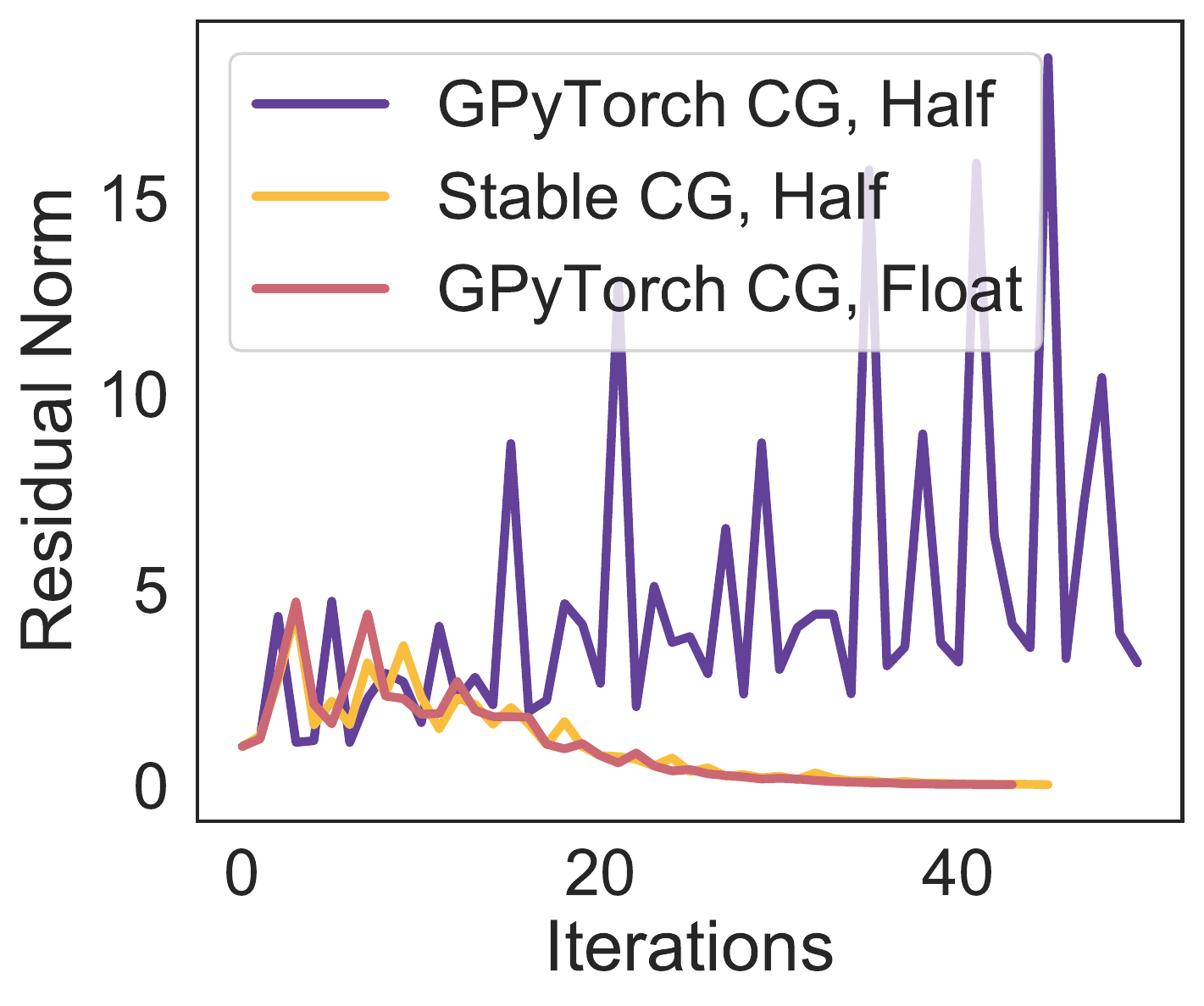}
			\caption{Buzz}
			\label{fig:buzz_types}
		\end{subfigure}
		\begin{subfigure}{0.23\textwidth}
			\centering
			\includegraphics[width=\linewidth]{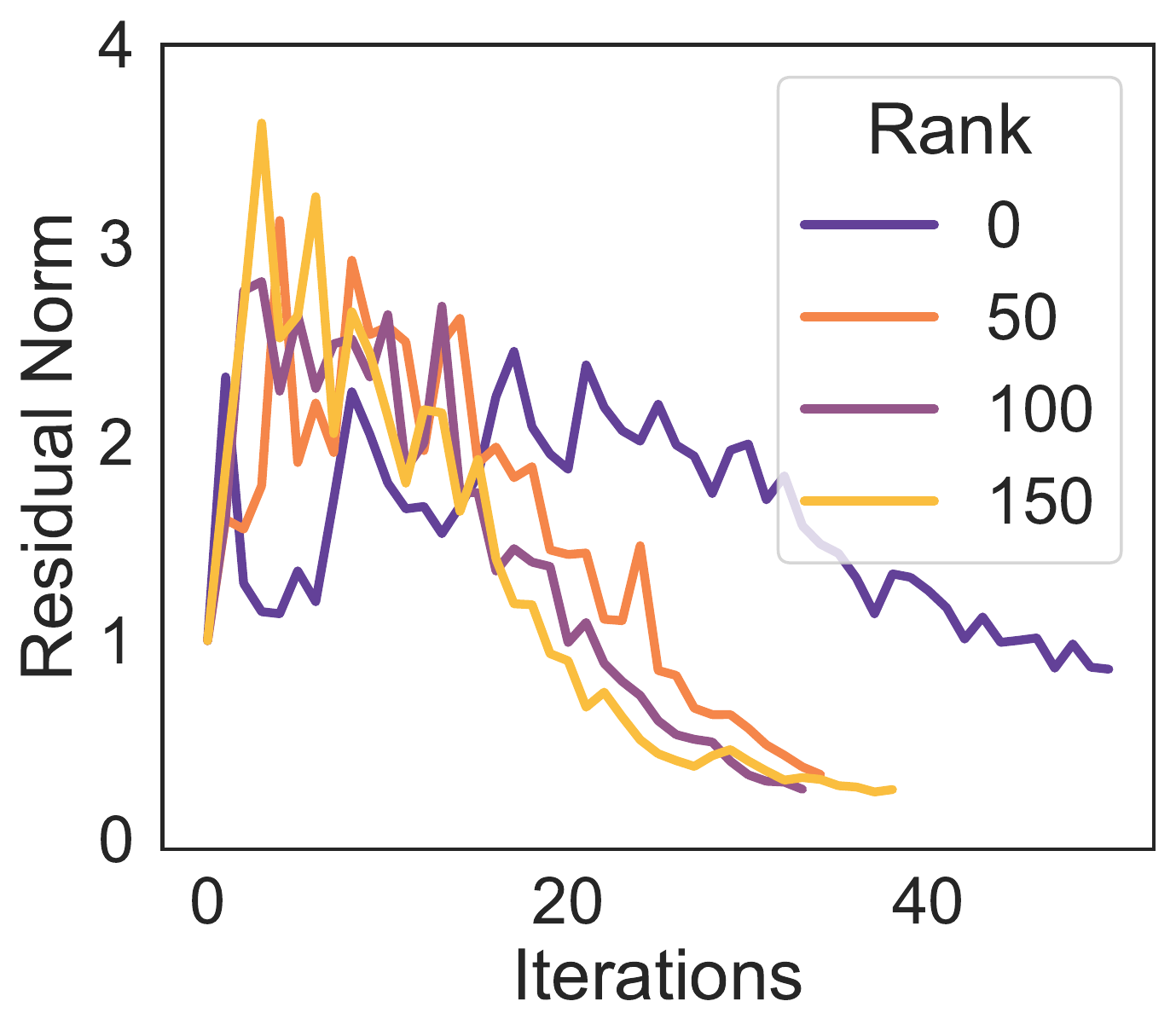}
			\caption{Elevators}
			\label{fig:elev_precond}
		\end{subfigure}
		\begin{subfigure}{0.23\textwidth}
			\includegraphics[width=\linewidth]{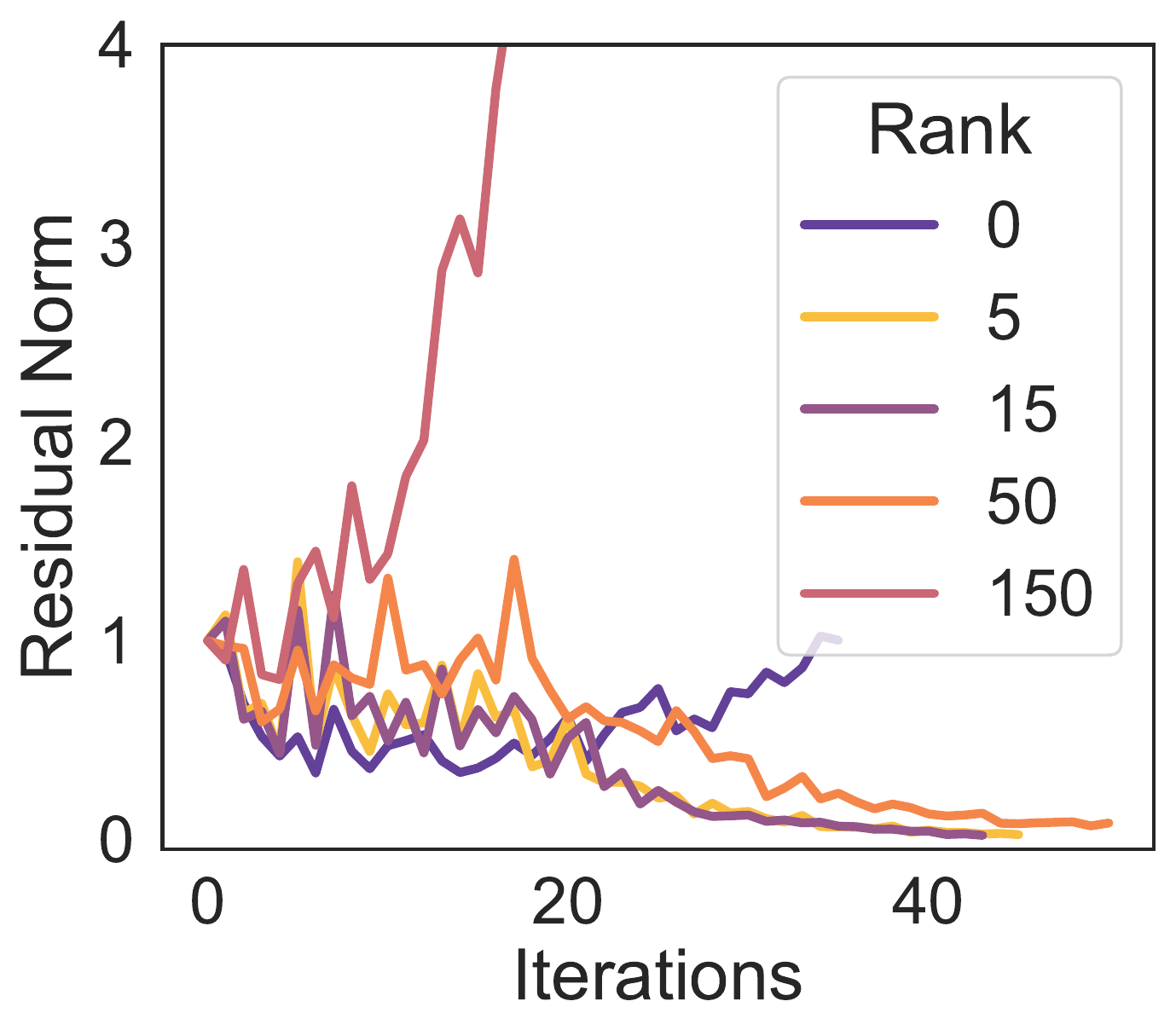}
			\caption{KeggD}
			\label{fig:kegg_precond}
		\end{subfigure}
		\begin{subfigure}{0.23\textwidth}
			\includegraphics[width=\linewidth]{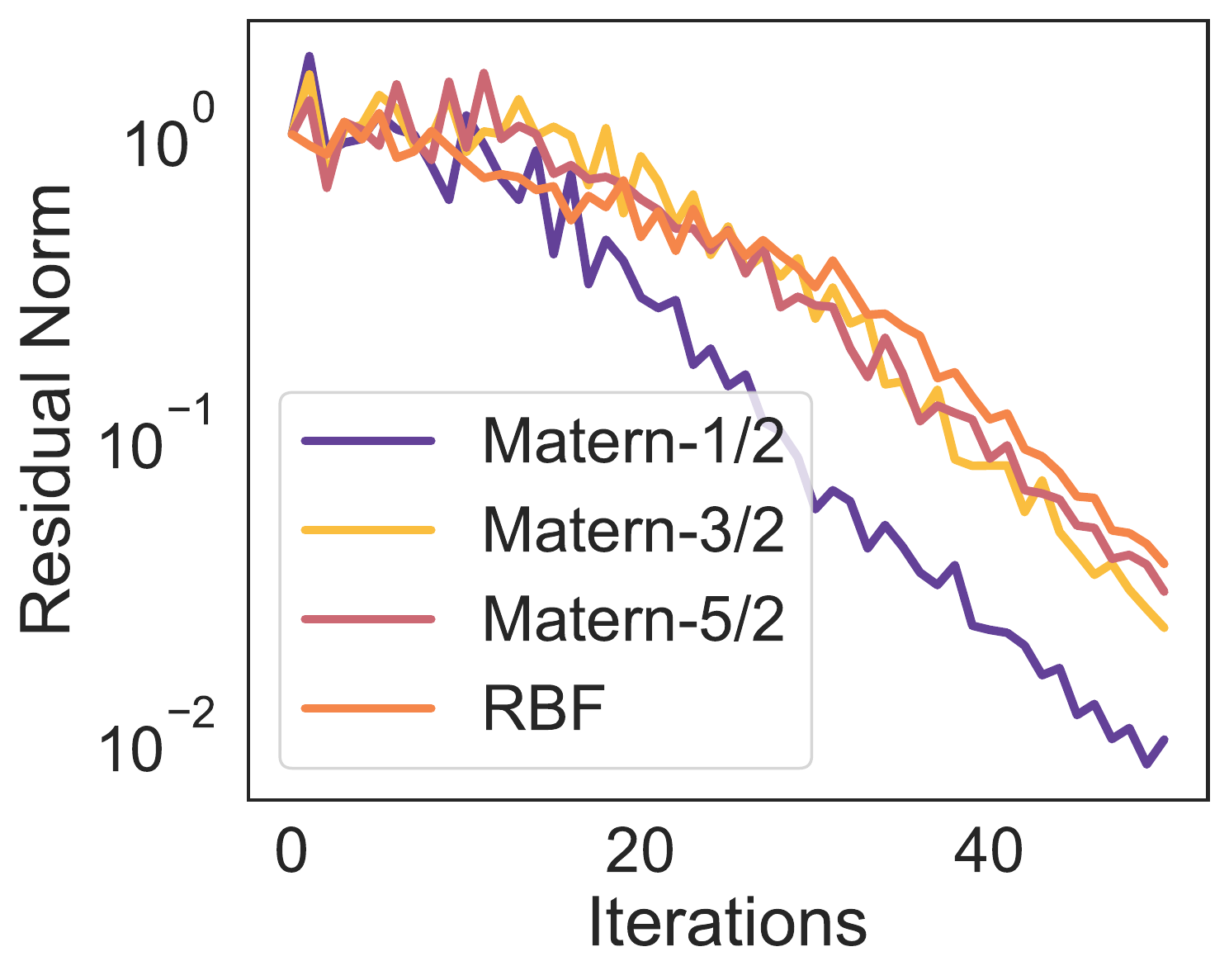}
			\caption{PoleTele}
			\label{fig:pol_precond}
		\end{subfigure}
		\caption{\textbf{(a)} Residual norms across solvers on buzz. \textbf{(b)} Residual norm for stable CG on elevators. Here no preconditioning also converges. \textbf{(c)} Residual norm for stable CG on Kegg Directed. \textbf{(d)} Residual norm for no ARD on Pol.}
		\label{fig:preconds_residual_tols}
	\end{figure}
	
	In Figure \ref{fig:buzz_types}, we display how CG in half diverges, but our stable CG converges as does CG in float.
	In Figure \ref{fig:elev_precond} and \ref{fig:kegg_precond}, we display the effect of preconditioning on solves, finding again that larger preconditioners tend to converge very slightly faster.
	
\begin{figure}[!ht]
		\centering
		\includegraphics[height=4cm]{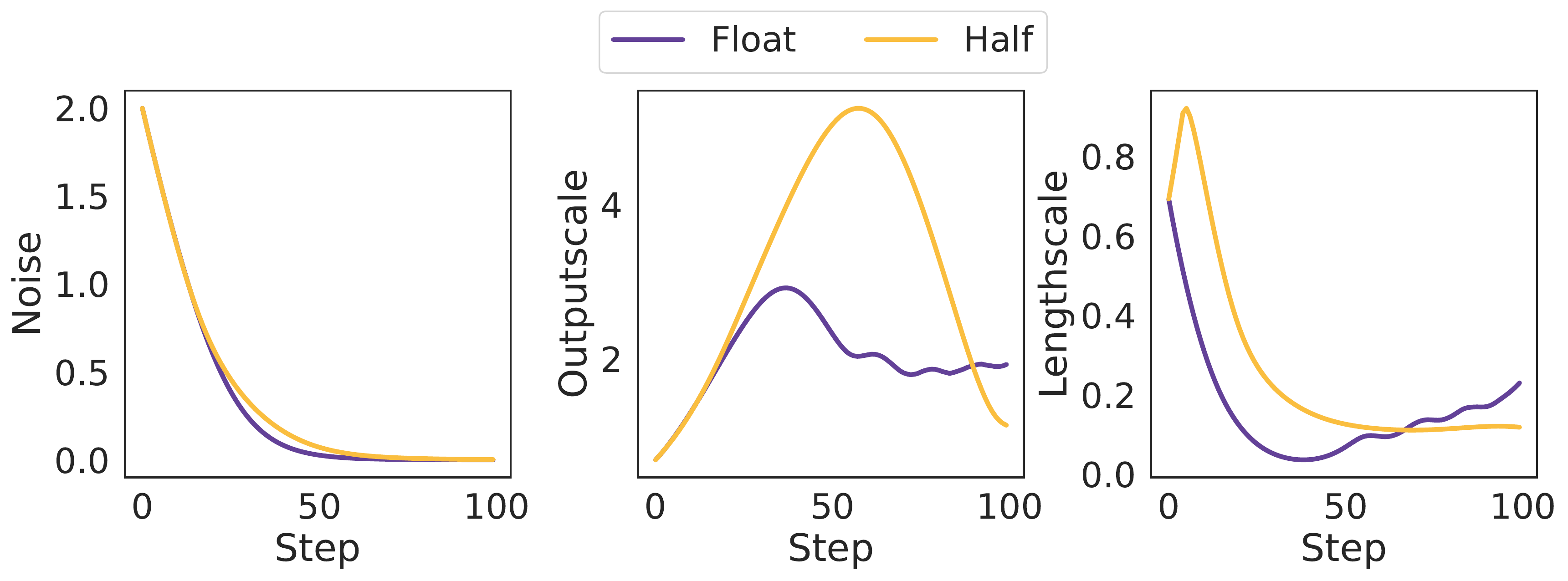}
		\caption{Optimization trajectory on \emph{3droad}. }
		\label{fig:3droad_comp}
	\end{figure}

	In Figure \ref{fig:3droad_comp}, we display the optimization trajectory on \emph{3droad} finding that there are clearer divergences in terms of the outputscale; however, each parameter converges to similar values by the end of training.
	
\begin{figure}[!ht]
      \centering
      \includegraphics[width=7cm,height=5cm]{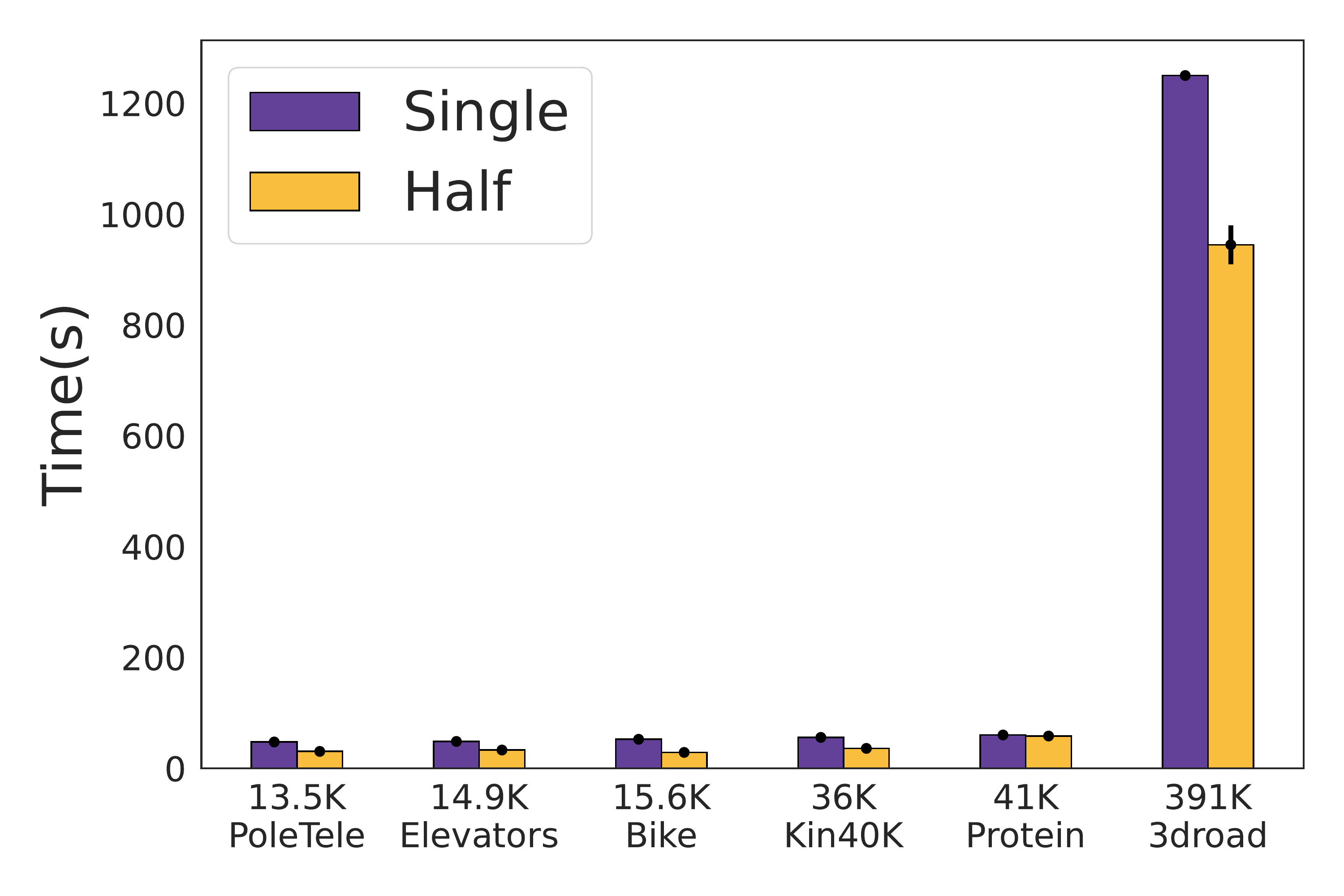}
      \includegraphics[width=7cm,height=5cm]{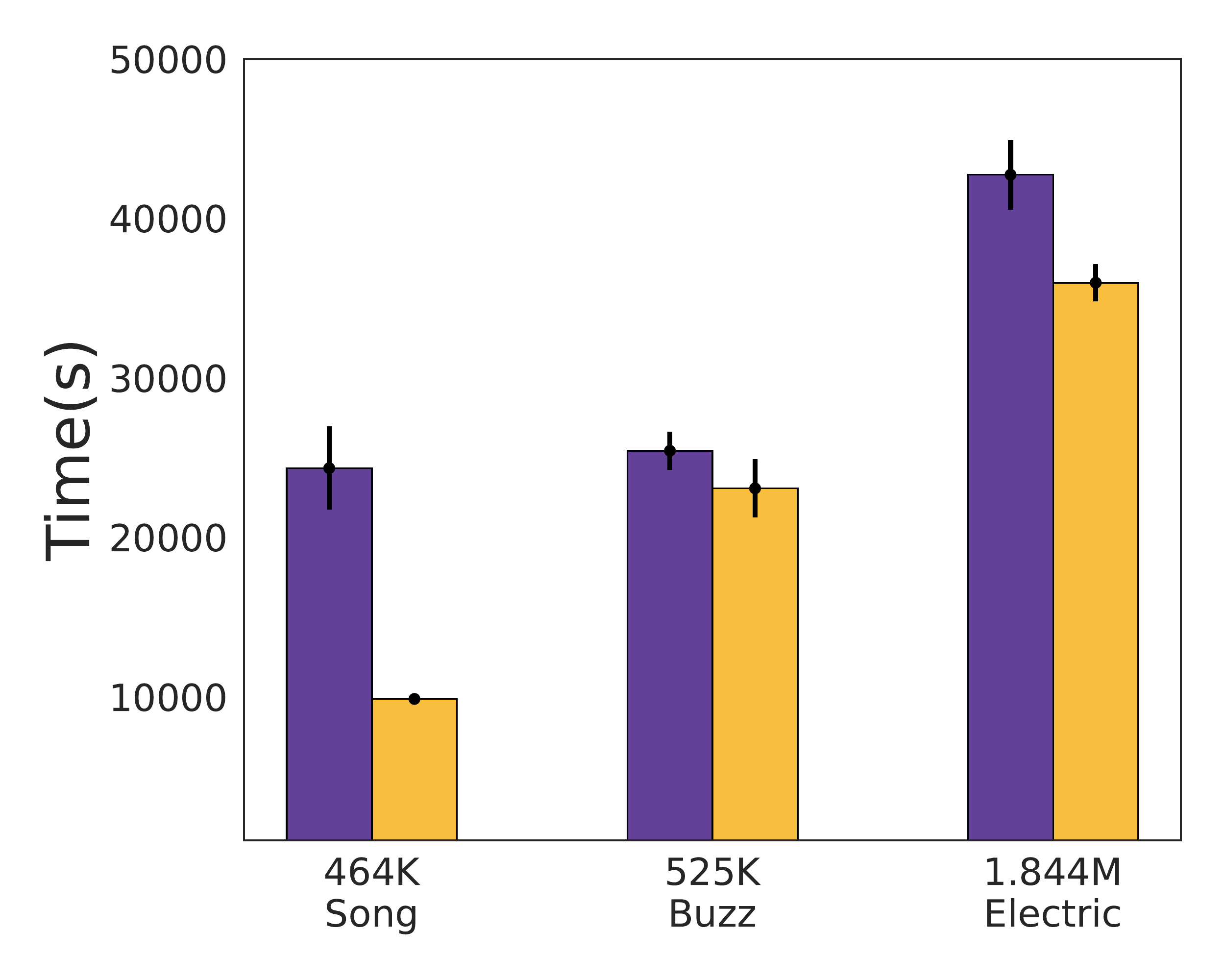}
      \caption{Running times without KeOps compilation for RBF kernel experiment on UCI.}
      \label{fig:without_comp}
    \end{figure}

\begin{figure}[!ht]
      \centering
      \includegraphics[scale=0.2]{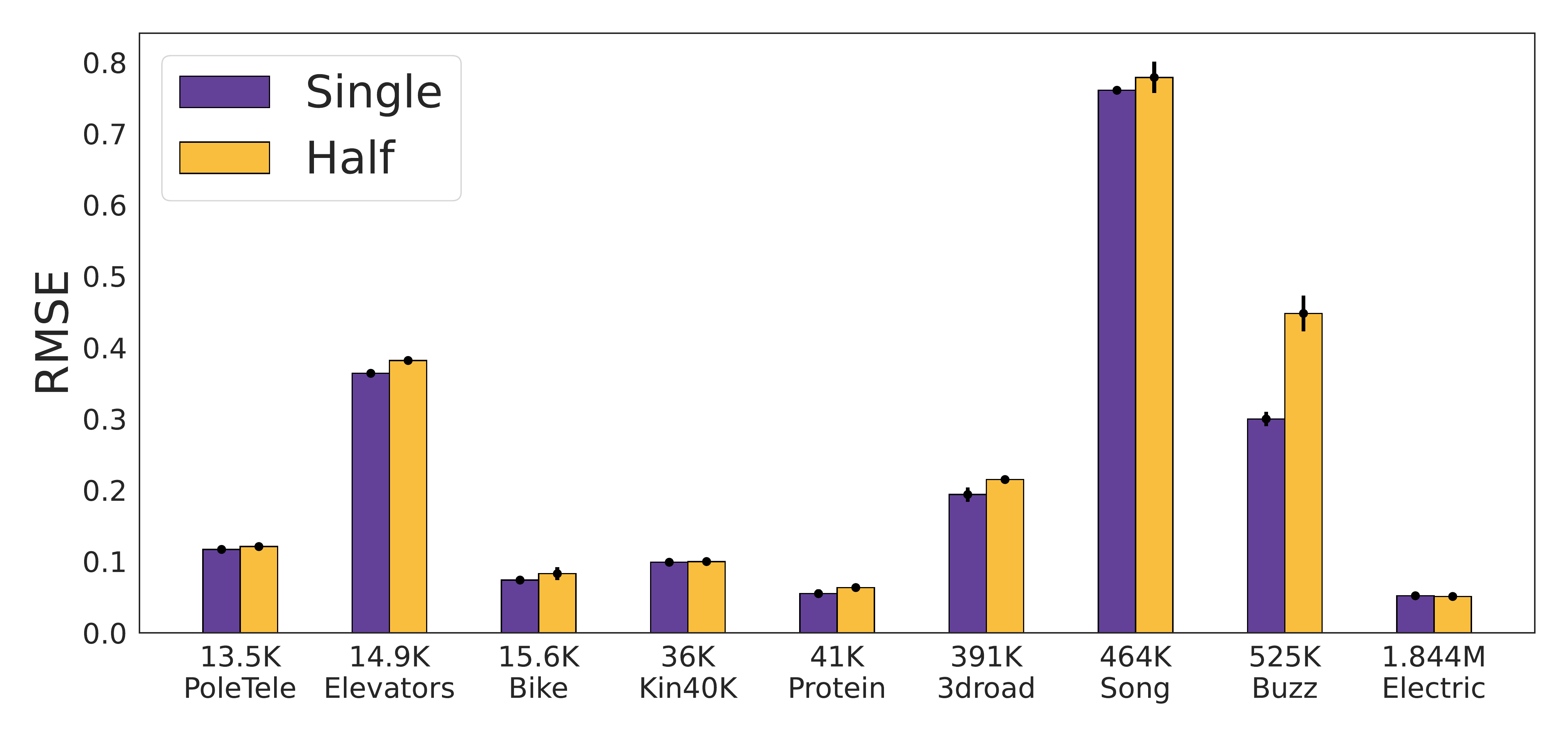}
      \caption{Comparison of RMSE for RBF kernel experiment on UCI.}
       \label{fig:rmse}
    \end{figure}

	\paragraph{Expanded Benchmark Results}
	In Figure \ref{fig:without_comp} we added the timing of Table \ref{tab:rbf_ard} without the compilation times of KeOps.
	Taking out the KeOps compilations times, half precision is always faster than single precision. 
	Additionally in Figure \ref{fig:rmse} we plotted the RMSE results where half precision, due to our stable method, maintains almost the same performance as
	single precision.
	
	\begin{figure}[!ht]
      \centering
      \includegraphics[width=7cm,height=5cm]{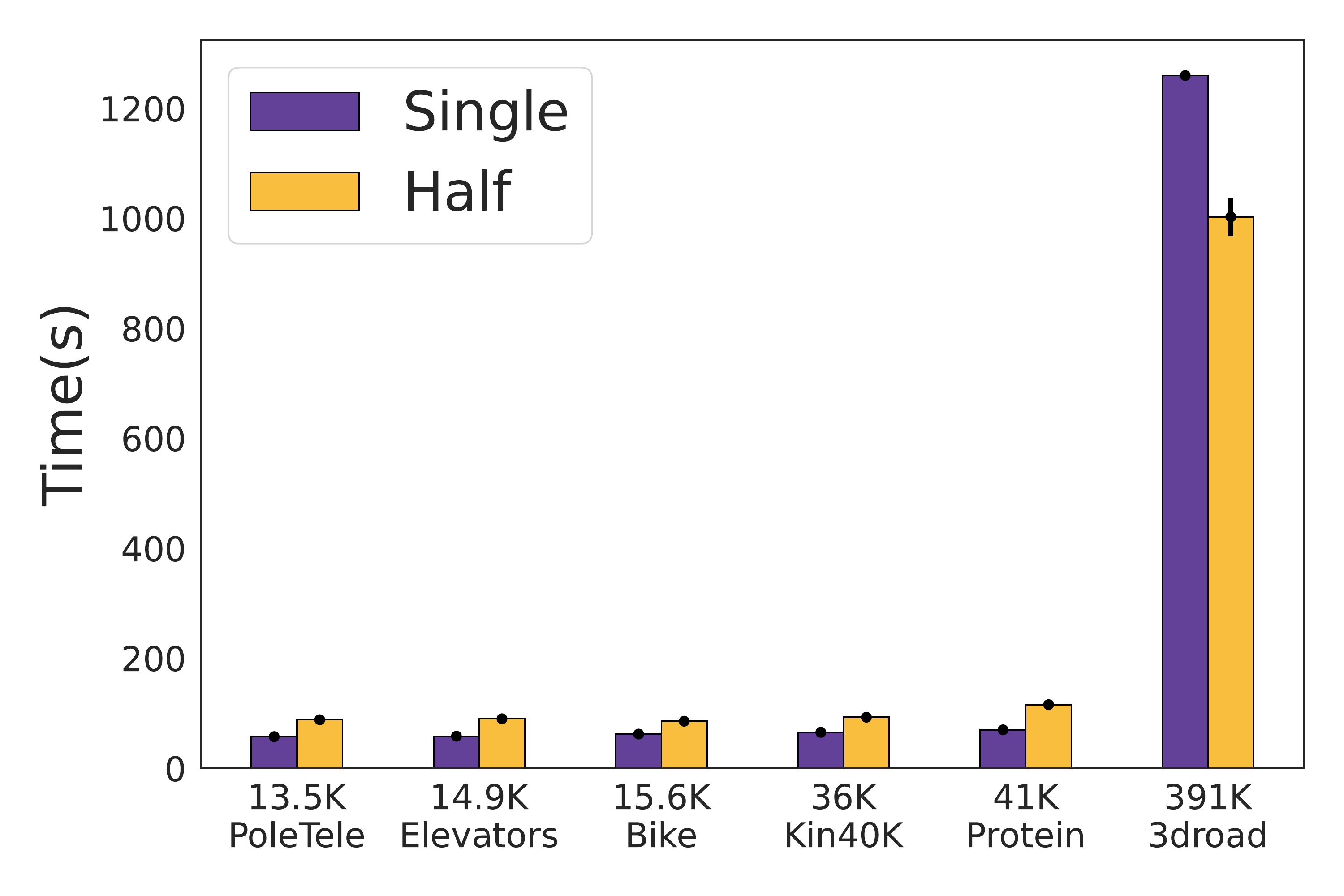}
      \includegraphics[width=7cm,height=5cm]{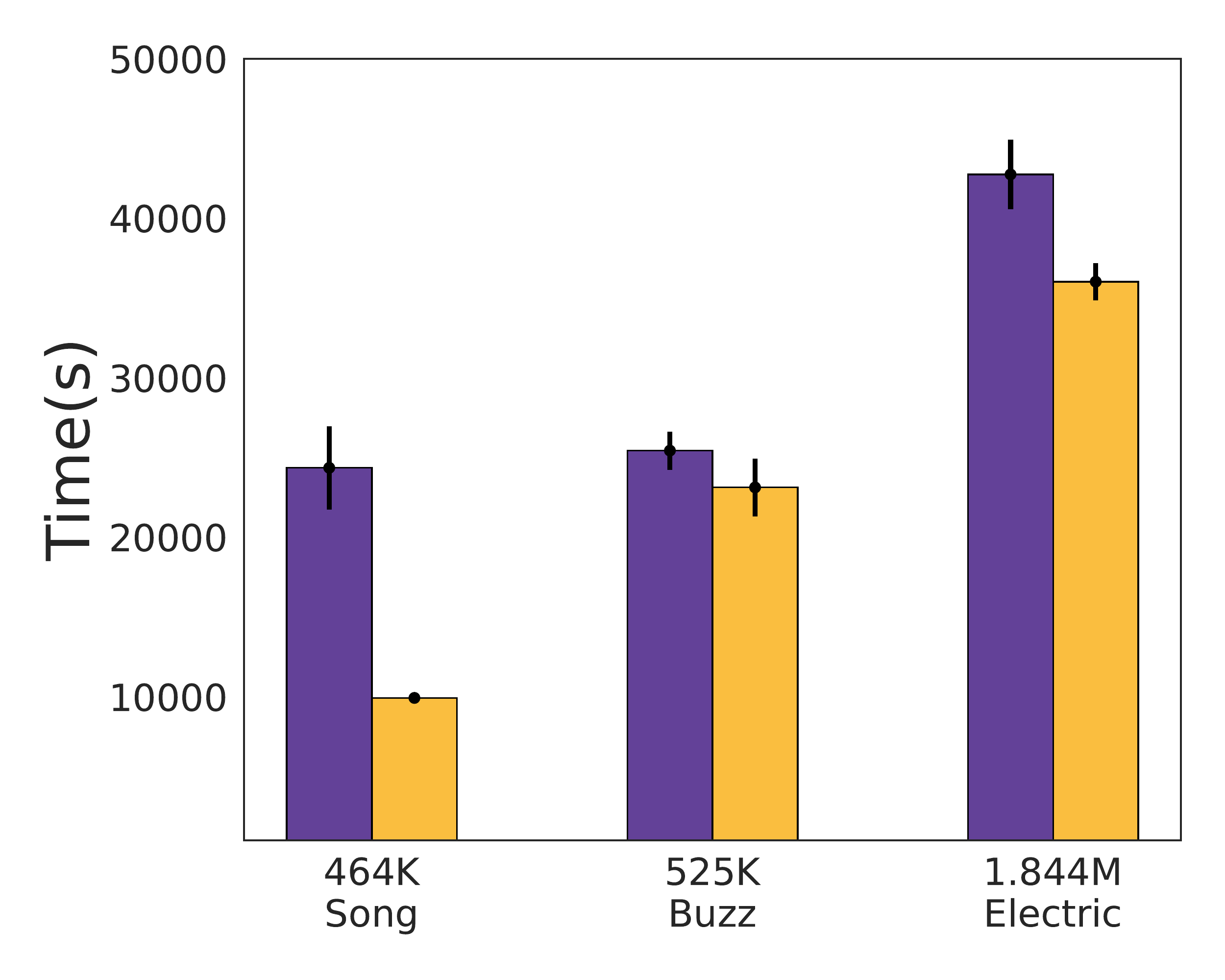}
      \caption{Running times with KeOps compilation for RBF kernel experiment on UCI.}
      \label{fig:with_comp}
    \end{figure}
	
	\paragraph{Additional Benchmark Results}
	
	In Table \ref{tab:rbf_nll}, we display NLLs across five seeds on UCI datasets for float, half, and SVGPs, analogous to our RMSE and timing results in Table \ref{tab:rbf_ard}.
	The timing results with and without PyKeops compilation times are shown in \ref{fig:with_comp} and \ref{fig:without_comp}, respectively.
	
	In Table \ref{tab:matern5_ard}, we display RMSEs, times, and NLLs for Mat\'ern-$5/2$ ARD kernels for both single and half precisions.
	In Table \ref{tab:matern3_ard}, we display RMSEs, times, and NLLs for Mat\'ern-$1/2$ ARD kernels for both single and half precisions. 
	In Table \ref{tab:matern1_ard}, we display RMSEs, times, and NLLs for Mat\'ern-$1/2$ ARD kernels for both single and half precisions.
	
	\begin{table}[!ht]
		\centering
		\caption{Test time NLLs across $5$ seeds on a suite of UCI tasks for float, half, and SVGPs with RBF ARD kernels.}
		\label{tab:rbf_nll}
		\begin{tabular}{ccccc}
			\midrule
			\multicolumn{1}{c}{\textbf{Dataset}} & \multicolumn{1}{c}{\textit{(N, d)}} & \multicolumn{1}{c}{\textbf{Single}} & \multicolumn{1}{c}{\textbf{Half}} & \multicolumn{1}{c}{\textbf{SVGP}} \\ 
			\hline
			PoleTele                    & (13.5K, 26)                & $-0.349 \pm 0.004$                 & $-0.316 \pm 0.004$               & $-0.513 \pm 0.011$               \\
			Elevators                   & (14.9K, 18)                & $0.515\pm 0.0195$                  & $0.663 \pm 0.024$                & $0.437 \pm 0.012$                \\
			Bike                        & (15.6K, 17)                & $-0.3714 \pm 0.0066$               & $-0.413 \pm 0.008$               & $-1.020 \pm 0.044$               \\
			Kin40K                      & (36K, 8)                   & $0.2352 \pm 0.005$                 & $0.241 \pm 0.005$                & $-0.327 \pm 0.007$               \\
			Protein                     & (41.1K, 9)                 & $0.9802 \pm 0.0115$               & $1.412 \pm 0.001$                & $0.964 \pm 0.015$                \\
			3droad                      & (391.4K, 3)                & $1.249 \pm 0.0129$                & $1.201 \pm 0.005$                & $0.537 \pm 0.025$                \\
			Song                        & (463.8K, 90)               & $1.146 \pm 0.0043$                 & $1.765 \pm 0.819$                & $1.418 \pm 0.002$                \\
			Buzz                        & (524.9K, 77)               & $-0.424 \pm 0.18$                  & $0.898 \pm 0.714$                & $-0.071 \pm 0.010$    \\
			Houseelectric & (1844.3K,9) & $-0.72 \pm 0.002$ & $-0.439 \pm 0.084$ & $-$
		\end{tabular}
	\end{table}
	
	\begin{table*}[!ht]
		\caption{RMSEs, NLL, and training time across $3$ seeds on a suite of UCI tasks. Here, we use Mat\'ern-$5/2$ ARD kernels with $50$ CG iterations and $50$ optimization steps.}
		\label{tab:matern5_ard}
		\centering
		\scriptsize{
			\begin{tabular}{ccccccccc}
				\midrule
				\multicolumn{2}{c}{\textbf{}} & 
				\multicolumn{2}{c}{\textbf{RMSE}} & 
				\multicolumn{2}{c}{\textbf{NLL}} & 
				\multicolumn{2}{c}{\textbf{Time}} & 
				\\
				\multicolumn{1}{c}{\textbf{Dataset}} & 
				\multicolumn{1}{c}{$(N , d)$} & 
				\multicolumn{1}{c}{\textbf{Single}} & 
				\multicolumn{1}{l}{\textbf{Half}} &
				\multicolumn{1}{c}{\textbf{Single}} & 
				\multicolumn{1}{l}{\textbf{Half}} &
				\multicolumn{1}{c}{\textbf{Single}} & 
				\multicolumn{1}{l}{\textbf{Half}} &
				\\ \hline
				PoleTele & (13.5K, 26) 
				&  $0.098 \pm 0.002$ & $0.102$ & $-0.454 \pm 0.004$ & $-0.447$ & $49.1 \pm 1.74$ & $97$\\
				Protein & (41.1K, 9) 
				& $0.498 \pm 0.007$ & $0.509 \pm 0.005$ & $1.07 \pm 0.01$ & $1.04 \pm 0.03$  & $69.17 \pm 1.7$ & $110 \pm 10$\\
				3droad & (391.4K, 3) 
				& $0.254 \pm 0.006$ & $0.231$ & $0.812 \pm 0.014$ & $0.899$ & $1666 \pm 214$ & $1501$\\
			\end{tabular}
		}
	\end{table*}
	
	\begin{table*}[!ht]
		\caption{RMSEs, NLL, and training time across $3$ seeds on a suite of UCI tasks. Here, we use Mat\'ern kernel $1/2$ with $50$ CG iterations and $50$ optimization steps.}
		\label{tab:matern1_ard}
		\centering
		\scriptsize{
			\begin{tabular}{cccccc}
				\midrule
				\multicolumn{2}{c}{\textbf{}} & 
				\multicolumn{1}{c}{\textbf{RMSE}} & 
				\multicolumn{1}{c}{\textbf{NLL}} & 
				\multicolumn{1}{c}{\textbf{Time}} & 
				\\
				\multicolumn{1}{c}{\textbf{Dataset}} & 
				\multicolumn{1}{c}{$(N , d)$} & 
				\multicolumn{1}{l}{\textbf{Half}} &
				\multicolumn{1}{l}{\textbf{Half}} &
				\multicolumn{1}{l}{\textbf{Half}} &
				\\ \hline
				PoleTele & (13.5K, 26) 
				& $0.108 \pm 0.003$ & $0.0478 \pm 0.006$ & $85 \pm 2.4$  \\
				Elevators & (14.9K, 18) 
				& $0.365 \pm 0.002$ & $-0.432 \pm 0.010$ & $83 \pm 1.9$  \\
				Bike & (15.6K, 17) 
				& $0.096 \pm 0.003$ & $0.527 \pm 0.004$ & $84 \pm 0.4$  \\
				Kin40K &(36K, 8) 
				& $0.136 \pm 0.002$ & $0.194 \pm 0.005$ & $87 \pm 0.1$\\
				Protein & (41.1K, 9) 
				& $0.481 \pm 0.003$ & $-0.762 \pm 0.013 $ & $98 \pm 0.1$ \\
				3droad & (391.4K, 3) 
				& $0.083 \pm 0.001$ & $-0.045 \pm 0.020$ & $1285 \pm 2.3$\\
			\end{tabular}
		}
	\end{table*}
	
	\begin{table*}[!ht]
		\caption{RMSEs, NLL, and training time across $3$ seeds on a suite of UCI tasks. Here, we use Mat\'ern kernel $3/2$ with $50$ CG iterations and $50$ optimization steps.}
		\label{tab:matern3_ard}
		\centering
		\scriptsize{
			\begin{tabular}{cccccc}
				\midrule
				\multicolumn{2}{c}{\textbf{}} & 
				\multicolumn{1}{c}{\textbf{RMSE}} & 
				\multicolumn{1}{c}{\textbf{NLL}} & 
				\multicolumn{1}{c}{\textbf{Time}} & 
				\\
				\multicolumn{1}{c}{\textbf{Dataset}} & 
				\multicolumn{1}{c}{$(N , d)$} & 
				\multicolumn{1}{l}{\textbf{Half}} &
				\multicolumn{1}{l}{\textbf{Half}} &
				\multicolumn{1}{l}{\textbf{Half}} &
				\\ \hline
				PoleTele & (13.5K, 26) 
				& $0.101 \pm 0.003$ &$0.475 \pm 0.005$ &$  90 \pm  2.2$ \\
				Elevators & (14.9K, 18) 
				& $0.498 \pm 0.112$ &$-0.990 \pm 0.423$ &$  86 \pm  3.5$ \\
				Bike & (15.6K, 17) 
				& $0.086 \pm 0.003$ &$0.567 \pm 0.006$ &$  88 \pm  0.1$ \\
				Kin40K &(36K, 8) 
				& $0.097 \pm 0.001$ &$0.186 \pm 0.004$ &$  96 \pm  0.1$ \\
				Protein & (41.1K, 9) 
				& $0.497 \pm 0.003$ &$-0.850 \pm 0.008$ &$ 107 \pm  0.1$ \\
				3droad & (391.4K, 3) 
				& $0.165 \pm 0.003$ &$-0.700 \pm 0.024$ &$1532 \pm  3.4$ \\
			\end{tabular}
		}
	\end{table*}
	
	\begin{table*}[!ht]
		\caption{RMSEs, NLL, and training time across $3$ seeds on a suite of UCI tasks. Here, we use Mat\'ern-$5/2$ kernels with $50$ CG iterations and $50$ optimization steps.}
		\label{tab:rbf_ard_app}
		\centering
		\scriptsize{
			\begin{tabular}{cccccc}
				\midrule
				\multicolumn{2}{c}{\textbf{}} & 
				\multicolumn{1}{c}{\textbf{RMSE}} & 
				\multicolumn{1}{c}{\textbf{NLL}} & 
				\multicolumn{1}{c}{\textbf{Time}} & 
				\\
				\multicolumn{1}{c}{\textbf{Dataset}} & 
				\multicolumn{1}{c}{$(N , d)$} & 
				\multicolumn{1}{l}{\textbf{Half}} &
				\multicolumn{1}{l}{\textbf{Half}} &
				\multicolumn{1}{l}{\textbf{Half}} &
				\\ \hline
				PoleTele & (13.5K, 26) 
				& $0.102 \pm 0.002$ &$0.437 \pm 0.011$ &$  96 \pm  2.1$ \\
				Elevators & (14.9K, 18) 
				& $0.520 \pm 0.209$ &$-0.967 \pm 0.571$ &$  94 \pm  5.9$ \\
				Bike & (15.6K, 17) 
				& $0.082 \pm 0.001$ &$0.568 \pm 0.008$ &$  93 \pm  0.1$ \\
				Kin40K &(36K, 8) 
				& $0.088 \pm 0.001$ &$0.078 \pm 0.002$ &$ 106 \pm  0.1$ \\
				Protein & (41.1K, 9) 
				& $0.512 \pm 0.008$ &$-1.048 \pm 0.008$ &$ 113 \pm 11.3$ \\
				3droad & (391.4K, 3) 
				& $0.226 \pm 0.005$ &$-0.905 \pm 0.007$ &$1486 \pm  4.5$ \\
			\end{tabular}
		}
	\end{table*}
	
	\section{Experimental Details}\label{sec:exp_details}
	
	\subsection{Maximum Distance Representable in finite precision}
	
	To compute these distances, we consider four separate stationary kernels \citep{rasmussen_gaussian_2008} with distance $d = |x  - x'|$ and lengthscale $l,$ focusing on determining what values they drop below a given $\epsilon.$
	For Mat\'ern-$1/2$ kernels, we need $\exp\{-d / l\} < \epsilon$ and solving gives $ d > -\log \epsilon * l$.
	For other Mat\'ern kernels (e.g. $3/2$ and $5/2$), there is no straightforward closed form solution, but empirical investigations showed that the maximum distance representable is somewhere between Mat\'ern-$1/2$ and RBF.
	
	For RBF Kernels, we have $ \exp\{- 1/2 d^2 / l^2\} < \epsilon$ and solving gets $d > (-2 \log \epsilon)^{1/2}  l$.
	For rational quadratic kernels, we have $(1 + \frac{d^2}{2 \alpha l^2})^{-\alpha} < \epsilon$
	and solving for $d$ gets $d > (2 \alpha (\epsilon^{1/\alpha} - 1))^{1/2} * l.$
	We found that for $\alpha = 2,3$ the size of the support was much larger, and so showed only $\alpha = 5.$
	For periodic kernels, we have
	$    -\frac{2}{\lambda}\sin^2\left(\frac{\pi}{p}|d|\right) < \log \epsilon$
	and solving gets
	$    |d| > \frac{p}{\pi}\arcsin\left(-\log \epsilon \frac{\lambda}{2}\right)$
	which will only have solutions when $-\log \epsilon \frac{\lambda}{2} \leq 1$.
	
	\subsection{Experimental Setup}
	
	All timing based experiments were performed using single NVIDIA V$100$ GPUs with $32$GB of memory on a shared supercomputing cluster. 
	Non-timing experiments also used NVIDIA RTX GPUs with either $24$ or $48$ GB of memory on either the same cluster or on a private internal server.
	
	We used GPyTorch \citep{gardner2018gpytorch} with the default parameter settings from Botorch's single task GP model\footnote{\url{https://botorch.org/api/models.html\#module-botorch.models.gp\_regression}.} which are constraining the noise to be greater than $0.0001$ and a Gamma$(1.1, 0.05)$ prior on the noise with initialization to $2$, and Gamma$(2.0, 0.15)$ prior on the outpuscale and a Gamma$(3.0, 6.0)$ prior on the lengthscale(s). 
	We fit using Adam for either $50$ or $100$ iterations unless otherwise documented and used a tolerance of $1.0$ for the CG iterations unless otherwise stated.
	We used KeOps $1.5$ for our experiments, noting that preliminary experiments with KeOps $2.0$ produced significantly faster compilation times \citep{charlier2021kernel}.
	
	For the datasets, we used the bayesian benchmarks package of \url{https://github.com/hughsalimbeni/bayesian_benchmarks/}, following their default training and testing splits.
	
	At test time, we converted the models back to float precision; however, our experiments found that this actually had limited impact on the RMSEs.
	
	\section{Theoretical Analysis} \label{app:theory}
	
	\subsection{Effect of finite precision on support}
	
	Following \citet[Chapter 4.3 of ][]{rasmussen_gaussian_2008} we can express the eigenvalues of an RBF kernel as 
	$\lambda_{k} = \sqrt{\frac{2 a}{A}} B ^{k}$ for some positive constants $a$, $A$ and $B \in \left(0, 1\right)$ that depend on the hyperparameters of the RBF kernel. 
	In infinite-precision an RBF kernel has support over the whole space as $\lambda_{k} > 0$. 
	However, if 
	\begin{equation*}
		\begin{split}
			k \geq \frac{\log \delta + \frac{1}{2}\log \left(\frac{A}{2a}\right)}{\log B} = \mathcal{O}(\log \delta)
		\end{split}
	\end{equation*}
	then $\lambda_{k} =0$ in finite-precision, where $\delta$ represents the round-off error. This means that the support of the kernel gets cut-off. This is similar to the support a Gaussian distribution $\mathcal{N}\left(\mu, \sigma^{2}\right)$ being the whole line $\mathbb{R}$, however, computing the probability of a sample being several standard deviations from the mean get cut-off to zero due to the sharp decay of the tails. Thus, our results focus on the empirical support of the kernel, not on the theoretical one.
	
	\subsection{Effect of finite-precision on generalization}
	Following \citet{li2019dimension}, we assume that moving from infinite precision to finite precision and using stochastic rounding means that our finite precision number, $Q(a) \sim U(a - \delta, a + \delta)$ for some $\delta$ that depends on our quantization (e.g. our precision based scheme). 
	Eqs. 11-15 of \citet{opper1998general} (see also Thm 1 of \citet{dicker2017kernel}, Thm 4.1 of \citet{zhang2005learning} and Thm 4 of \citet{caponnetto_optimal_2007}), generalization bounds often depend on the effective dimension of the training kernel matrix. Recall that the effective dimension is computed from the eignenvalues as $\sum_{i=1}^N \frac{\lambda_i}{\lambda_i + s}$ for some value $s$. 
	We will compute our finite precision approximation by computing the expected value of the effective dimension under the stochastic rounding scheme.
	
	Furthermore, we assume that each eigenvalue is quantized independently, to compute the expected effective dimension, we need to compute $N$ integrals of the following form, where $p(x) = U(a - \delta, a + \delta)$:
	\begin{align}
		\mathbb{E}_{p(x)}\frac{x}{x+s} = \int_{a-\delta}^{a + \delta} &\frac{x}{x+s} \frac{1}{a + \delta - (a - \delta)} dx = \frac{1}{2\delta} (x - s \log (x + s))|_{a-\delta}^{a + \delta} \nonumber \\
		&=\frac{1}{2\delta}\left(a + \delta - s \log(a + \delta + s) - (a - \delta - s \log(a - \delta + s)\right) \nonumber \\
		&=1 + \frac{s}{2 \delta}\log\frac{a + s - \delta}{a + s + \delta} = 1 + \frac{s}{2 \delta}\log\left(1 - \frac{2\delta}{a + s + \delta}\right) \nonumber \\
		&=1 - \frac{s}{2\delta}\left(\frac{2\delta}{a + s + \delta} + \frac{4\delta^2}{2 (a + s + \delta)^2} + \frac{8\delta^3}{3 (a + s + \delta)^3} + \mathcal{O}(\delta^4)\right) \nonumber \\
		&=1 - \frac{s}{a + s + \delta} - \frac{\delta s}{(a + s + \delta)^2} - \frac{4 \delta^2 s}{3(a + s + \delta)^3} - \mathcal{O}(\delta^3) \\
		&\geq 1 - \frac{s}{a + s + \delta} - \frac{\delta s}{(a + s + \delta)^2}
	\end{align}
	Now, putting this into our expectation over the quantized eigenvalues:
	\begin{align}
		\mathbb{E} \left( \sum_{i=1}^N \frac{Q( \lambda_i)}{ Q(\lambda_i) + s} \right) &\geq \sum_{i=1}^N 1 - \frac{s}{\lambda_i + s + \delta} - \frac{\delta s}{(\lambda_i + s + \delta)^2} \\
		&= N - \sum_{i=1}^N \frac{s(\lambda_i + s + \delta) - \delta s}{(\lambda_i + s + \delta)^2} = N - \sum_{i=1}^N \frac{s(\lambda_i + s)  }{(\lambda_i + s + \delta)^2} \nonumber \\
		&\geq N - \sum_{i=1}^N \frac{s}{\lambda_i + s} \nonumber \\
		&=N_{\text{eff}}(K, s) \nonumber
	\end{align}
	Note that as $\delta \rightarrow 0$ all of these inequalities become tight as expected. What this shows is that in finite precision, the expected effective dimensionality can only be higher than the effective dimensionality in infinite precision.
	
	In general, bounds such as Eqs. 11, 16 of \citet{opper1998general} (also Eq. 7.26 in \citet{rasmussen_gaussian_2008} tend to depend on the expected eigenvalues rather than those estimated empirically (e.g. in finite precision). However, they are related as $\frac{1}{N} \lambda_i^{\text{emp}} \rightarrow \lambda_i$ (see Rasmussen \& Williams, 06 4.3.2) and so we estimate $N_{\text{eff}}(K, \sigma^2 / n)$ with $N_{\text{eff}}(K^{\text{emp}}, \sigma^2)$. 
	Plugging in our finite precision estimate to something like Thm. 4.1 of \citet{zhang2005learning} then suggests that the generalization error in (any) finite precision will tend to be higher than for infinite precision.
	Roughly, these bounds state that the generalization error of a kernel ridge regressor is upper bounded by the sum of approximation error terms (relating to the fit of the kernel to the function) plus $N_{\text{eff}}(K, \lambda) / n$ for a regularization term $\lambda$ (similar to the noise value in the GP setting).
	
	Finally, our analysis shows that a larger $\delta$ (e.g. a lower precision estimate) will tend to further increase the generalization error.
	This tends to confirm our experimental study on the effective dimension in Figure \ref{fig:matern_spectrum}.

\end{document}